\documentclass{article}
\PassOptionsToPackage{numbers, compress}{natbib}

\usepackage[main, final]{neurips_2026}
\usepackage{adjustbox}

\usepackage[utf8]{inputenc} 
\usepackage[T1]{fontenc}    
\usepackage{hyperref}       
\usepackage{url}            
\usepackage{booktabs}       
\usepackage{amsfonts}       
\usepackage{nicefrac}       
\usepackage{microtype}      
\usepackage{xcolor}         
\usepackage{rotating}       

\usepackage{amsmath}
\usepackage{amssymb}
\usepackage{booktabs}
\usepackage{tabularx}
\usepackage{array}
\usepackage{pifont}
\usepackage{graphicx}
\usepackage{enumitem}
\usepackage{makecell}
\usepackage{hyperref}

\newcolumntype{L}{>{\raggedright\arraybackslash}X}

\title{MC$^{2}$: Monte Carlo Correction for Fast Elliptic PDE Solving}

\author{%
  Ethan Hsu\thanks{Equal contribution}%
  \kern0.5em\thanks{Corresponding author}\\
  Stanford University\\
  \texttt{contact@ethanhsu.net}\\
  \texttt{ethanhsu@stanford.edu}
  \And
  Ivan Ge\footnotemark[1]\\
  Stanford University\\
  \texttt{ivange@stanford.edu}
  \And
  Hong Meng Yam\footnotemark[1]\\
  Stanford University \\ Pinetree Research\\
  \texttt{yam@pinetree-research.com}
}

\begin{document}

\maketitle



\begin{abstract}
    
Partial differential equation (PDE) solvers underpin modern computer graphics, geometry processing, and scientific computing. Classical Monte Carlo PDE solvers such as Walk-on-Spheres (WoS) provide mesh-free estimations that are unbiased, but with accuracy limited by sampling and numerical approximation. Learned solvers are fast but biased and brittle under distribution shift. We therefore present \textbf{MC$^2$}, a hybrid Neural Network-Monte Carlo PDE solver that treats a low-budget Monte Carlo solution as a structured estimator of the true field and learns a single-pass neural correction to recover a high-fidelity solution. MC$^2$ matches the accuracy of solutions using over $1000\times$ more Monte Carlo samples, outperforming all evaluated mesh-free classical, denoising, and neural-operator baselines. To enable reproducible study of finite-compute PDE solving, we additionally release \textbf{PDEZoo}, the largest standardized elliptic PDE benchmark to date: 2 million PDEs spanning 5 elliptic families and unlimited geometric compositions, with analytic ground truth and multi-budget Monte Carlo trajectories. Together \textbf{MC$^2$} and \textbf{PDEZoo} (1) empirically establish that finite-sample Monte Carlo error is structured, learnable, and correctable in a single forward pass, (2) show that we can solve PDEs $\sim$\textbf{218x} faster than WoS at matched accuracy and a $\sim$ \textbf{171x} reduction in inference time relative to SOTA diffusion methods with higher accuracy, and (3) provide the reproducible evaluation infrastructure the field has so far lacked.
\href{https://mc-error-correction.github.io}{Code \& Data: https://mc-error-correction.github.io}

\end{abstract}







\section{Introduction}
\begin{figure}
    \centering
    \includegraphics[width=1\linewidth]{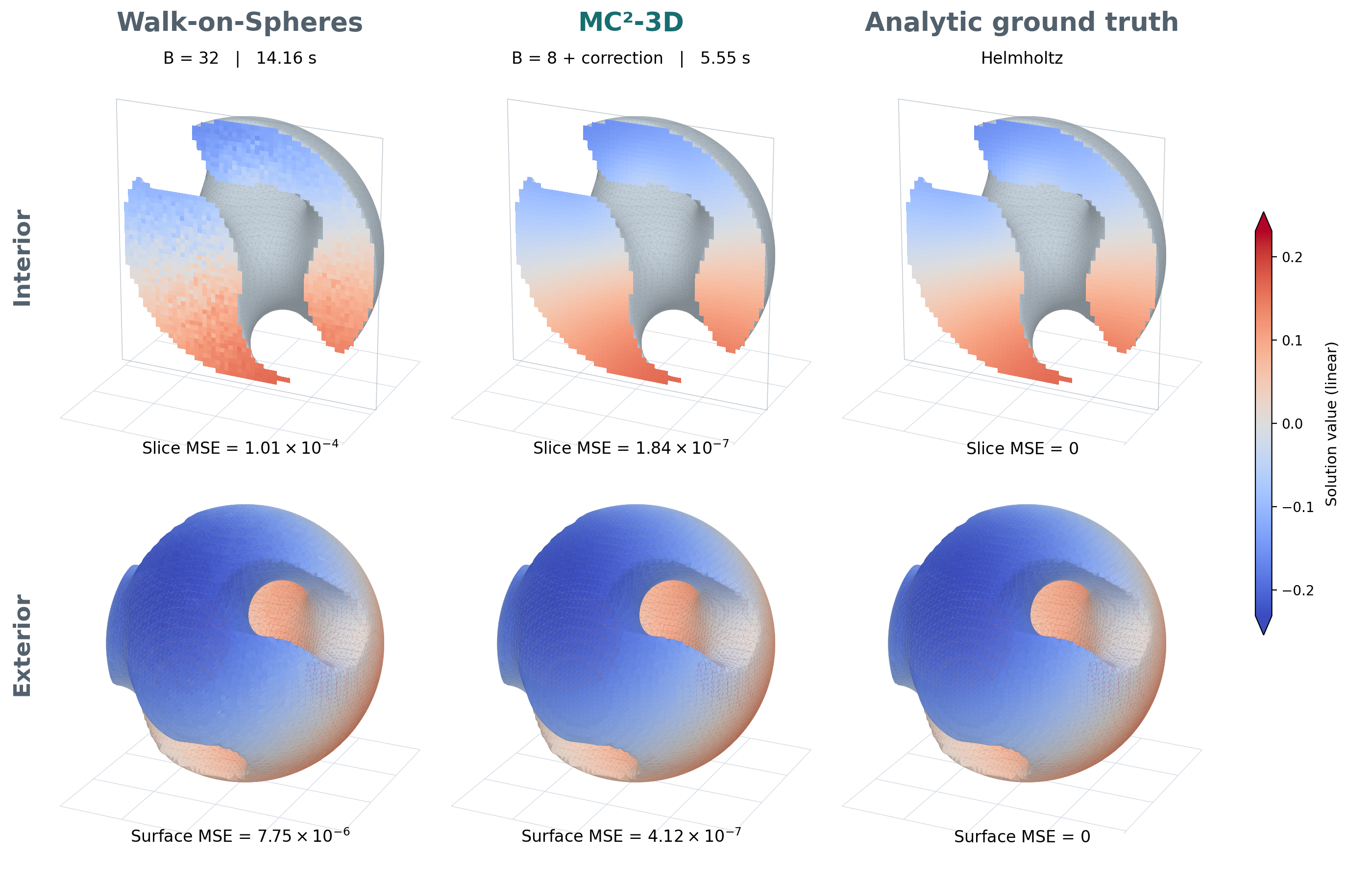}
    \caption{Three-dimensional Helmholtz solutions on a gyroid domain. Interior slices (top) and exterior views (bottom) compare WoS with $B=32$, MC$^2$ with $B=8$, and analytic ground truth using a shared linear color scale. MC$^2$ closely recovers the solution structure while suppressing the sampling noise visible in WoS, reducing interior-slice MSE by approximately $550\times$. MSE is reported below each panel.}
  \label{fig:3d_gyroid_comparison}
\end{figure}

Elliptic partial differential equations (PDEs) are a core computational primitive across scientific computing, geometry processing, and computer graphics, critical for tasks such as reconstruction, simulation, and physical modeling \citep{botsch2010polygon, crane2013digital, NEURIPS2022_93476ae4}. In practice, solvers must operate reliably across diverse forcing functions, boundary conditions, and geometries, while remaining fast enough for large-scale or latency-sensitive pipelines.  

Classical deterministic solvers such as finite difference and finite element methods (FDM/FEM) can be highly accurate, but rely on explicit meshing and problem-specific discretization choices that limits flexibility on irregular domains \cite{briggs2000multigrid, brenner2008mathematical, Sawhney:2020:MCG}. Monte Carlo solvers, most notably Walk-on-Spheres (WoS), address many of these limitations. Their ideal representations are mesh-free and unbiased under their assumptions, and apply to broad classes of elliptic PDEs and geometries \cite{Muller, Sawhney:2020:MCG, 10.1145/3528223.3530134}. However, their finite-iteration outputs suffer from high variance, requiring orders of magnitude more samples to reach acceptable accuracy \cite{Sawhney:2023:WoSt}. 

Learning-based solvers, such as PINNs, neural operators, transformer models, offer fast inference but introduce bias, and can fail under distribution shifts or unseen PDE specifications \cite{berg2018unified, khoo2021solving, cuomo2022scientific, 10.1145/3610548.3618141}. Recent hybrid approaches integrate neural networks into Monte Carlo solvers such as neural caches \cite{zhang2023monte}) or condition generative models on sparse numerical samples \cite{huang2024diffusionpde}, with training and inference requirements that vary by method but can suffer from per-problem retraining or slow iterative inference.


In this work, we propose MC²: a new approach to hybrid PDE solving that treats the low-iteration output of a Monte Carlo solver as a structured estimator of the true solution and learns a neural network correction that reduces error in a single forward pass. Rather than modifying a Monte Carlo solver or replacing it with a learned surrogate, MC² applies a transformer-based field-to-field model to early Walk-on-Spheres outputs, recovering solution quality measured by PSNR that would otherwise require around 1000× more Monte Carlo iterations \cite{Fan_2022}. We show that this preserves the generality and robustness of Monte Carlo solvers while dramatically accelerating convergence and lowering the bias of standalone neural solvers. While the learned correction can indeed introduce bias, we show that empirical evaluations demonstrate the maintained accuracy and robustness. 


To support this setting and enable reproducible evaluation, we introduce \textbf{PDEZoo}, a benchmark offering varied two-dimensional, constant-coefficient elliptic problems. PDEZoo spans Laplace, Poisson, Yukawa, Biharmonic, and Helmholtz equations with explicit forcing functions, boundary conditions, and dense reference solutions. PDEZoo is designed for finite-compute evaluation, it includes Monte Carlo solver trajectories at multiple compute budgets, exposing how solvers behave before asymptotic convergence and allowing methods to be compared under practical resource constraints. We further release the full procedural generator, enabling arbitrarily large training sets while preserving a fixed benchmark for standardized evaluation.

Our contributions are twofold: 
\begin{enumerate}
\item \textbf{MC²}, a hybrid WoS–NN solver that corrects structured Monte Carlo error in WoS fields across diverse elliptic PDEs, matching the quality of using $\sim$1000× more Monte Carlo samples per pixel and outperforming all evaluated baselines. Therefore, it combines the generality and mesh-free capabilities of Monte Carlo PDE solvers with learned error correction to yield a more than 200$\times$ wall-clock speedup on PDE solving without sacrificing accuracy.
\item \textbf{PDEZoo}, the large-scale standardized benchmark for elliptic PDEs with explicit forcing and boundary functions, augmented with Monte Carlo solver trajectories and a scalable procedural generator. 

\end{enumerate}

\section{Related Work}  

\textbf{Monte Carlo PDE Solvers.} Walk-on-Spheres (WoS) estimates solutions to elliptic PDEs via random walks guided by Green's functions \cite{Muller}. Ideal WoS representations are unbiased under their assumptions, whereas practical shell termination and walk truncation can introduce bias. WoS is mesh-free, extends to irregular geometries, and has been generalized to mixed Dirichlet--Neumann conditions \cite{Sawhney:2023:WoSt}, boundary integral formulations \cite{Sugimoto_2023}, and Monte Carlo finite element settings \cite{Jin}. These properties make WoS attractive for graphics and simulation \cite{Sawhney:2020:MCG, 10.1145/3528223.3530134}. For independent walks with finite variance, the average has variance $\mathcal{O}(1/K)$ and standard deviation $\mathcal{O}(1/\sqrt{K})$. As a result, reaching acceptable accuracy in budget-constrained settings can require prohibitively many samples.

\textbf{Neural Networks for PDEs.} Physics-informed neural networks (PINNs) enforce PDE residuals through specialized loss functions \cite{berg2018unified, jiang2023neural, raissi2017physics}, while neural operators such as FNO learn mappings between function spaces \cite{li2021fourier}. More recent work has explored generative approaches using diffusion models. DiffusionPDE conditions a diffusion model on sparse observations to jointly recover coefficients and solutions while CoCoGen augments score-based sampling with PDE-residual guidance to improve physical consistency \cite{ huang2024diffusionpde, jacobsen2024cocogen}. Transformer-based architectures have further enabled fast inference and some degree of cross-PDE generalization. However, neural solvers face non-convex loss landscapes, spectral bias \cite{DBLP:journals/corr/abs-2007-14527, rathore2024challengestrainingpinnsloss}, and distribution-shift fragility, producing biased solutions that degrade on out-of-distribution problems \cite{10.1145/3610548.3618141, cuomo2022scientific}.

\textbf{Hybrid Monte Carlo-Neural Network Methods.} Recent work combines neural networks with Monte Carlo solvers, trading approximation bias for reduced variance and computational cost. \citet{10.1145/3610548.3618141} trains a neural field to approximate the PDE solution, then use it as a control-variate cache during WoS sampling. While effective, this requires per-problem training and manual hyperparameter tuning. Diffusion-based approaches condition generative models on sparse numerical samples to reconstruct full solution fields \cite{huang2024diffusionpde}, but suffer from slow multi-step inference and conditioning instability \cite{kawar2022denoising}. Other methods such as Boundary Value Caching (BVC) uses a per-problem boundary cache without neural training, reusing boundary integrals without added statistical bias \cite{miller2023bvc}. Neural Control Variates with Automatic Integration (NCV-AI) learns integrable control variates and retains an unbiased Monte Carlo integral estimator \cite{li2024ncvai}. Neural Walk-on-Spheres (NWoS) trains a neural solution using WoS losses \cite{nam2024nwos}, while Walk-on-Spheres Neural Operator (WoS-NO) trains across PDE instances and predicts directly \cite{viswanath2026wosno}. Our approach differs in two key ways: (i) the corrector is trained once across the entire PDE distribution rather than per-instance, and (ii) correction is a single deterministic forward pass, not iterative sampling.

\textbf{Monte Carlo Denoising in Rendering.}
Similar ideas of feature-guided filtering \cite{rousselle2013robust}, kernel-predicting networks \cite{bako2017kpcn}, recurrent autoencoders \cite{chaitanya2017recurrent}, and sample-based kernel-splatting networks \cite{gharbi2019samplebased} are used in rendering to reconstruct low-sample Monte Carlo estimates. Our contribution mainly focuses on the WoS/elliptic-PDE instantiation, source-conditioned correction across PDE instances, and evaluation infrastructure with multi-budget trajectories.

\textbf{Benchmarks and Scientific Progress.}
Despite a rapidly growing literature on neural and hybrid PDE solvers, the field lacks the unified, standardized evaluation infrastructure that has driven progress in adjacent areas of machine learning like NLP or computer vision, such as GLUE \cite{wang2018glue}, SuperGLUE \cite{wang2019superglue}, and ImageNet \cite{deng2009imagenet}. Existing benchmarks each occupy narrow regimes: PDEBench \cite{takamoto2022pdebench} and PDEArena \cite{gupta2022towards} target time-dependent PDEs on fixed rectangular grids, the FNO datasets \cite{li2021fourier} ship a small number of canonical problems on the unit square, and Poseidon \cite{herde2024poseidon}, while large-scale, is restricted to fixed grids and uses numerical solver outputs as ground truth. Crucially, none of these benchmarks support \emph{finite-compute} evaluation. Rather, they provide a single converged solution per instance, with no visibility into solver behavior at intermediate sample budgets. They also lack analytic ground truth, conflating model error with discretization error from the reference solver. Furthermore, they all lack a maintained leaderboard and sufficient test sizes for the community to accurately track State-Of-The-Art results. As a result, comparisons across papers in this space are typically apples-to-oranges, and there is no shared standard against which new methods are evaluated. We designed PDEZoo to fill this gap.

\section{Methodology: Monte Carlo Error Correction}


We introduce a framework for mesh-free, fast elliptic PDE solving in the finite-compute regime.

\begin{figure}
        \centering
        \includegraphics[width=0.75\linewidth]{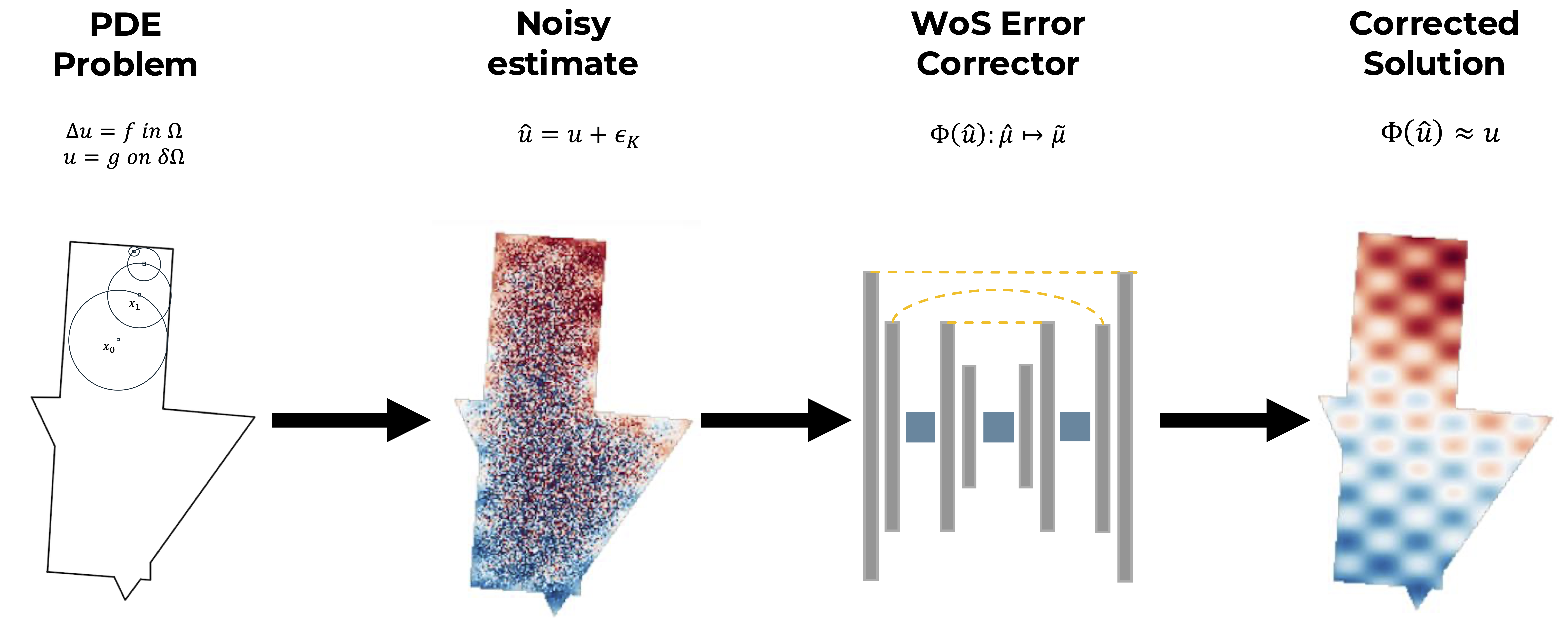}
        \caption{\textbf{MC$^2$ pipeline.} The headline corrector receives a low-budget WoS field and its source term as two input channels and predicts a corrected field in one forward pass.}
        \label{fig:mc2_pipeline}
\end{figure}
\textbf{Problem Setting and PDE Families} The second-order benchmark families use constant coefficients and satisfy:
\vspace{-0.25\baselineskip}
\begin{equation}
  \begin{cases}
    \Delta u(x)-c u(x)=f(x), & x\in\Omega \\
    u(x)=g(x), & x\in\partial\Omega
  \end{cases}
\end{equation}
Here $c=0$ for Laplace/Poisson, $c=\lambda>0$ for Yukawa, and $c=-k^2$ for Helmholtz.  The forcing function is $f$, the boundary function is $g$, and $\Omega$ is specified via a signed distance function \cite{10.1145/3610548.3618141}. 

\textbf{Monte Carlo PDE Solvers as Stochastic Estimators.} Monte Carlo solvers such as Walk-on-Spheres (WoS) estimate solutions using random walk payoffs derived from Green's functions. Let $Y_j(x)$ denote a complete walk payoff, including boundary and accumulated source contributions, with numerical settings $\eta$ suppressed. For a fixed PDE instance $\mathcal P$, independent identically distributed payoffs with finite variance give $\hat u_K=K^{-1}\sum_{j=1}^K Y_j$, with $\mathbb E[\hat u_K\mid\mathcal P]=u+b_\eta$ and $\operatorname{Var}(\hat u_K\mid\mathcal P)=\operatorname{Var}(Y_1\mid\mathcal P)/K$. Sampling variance can dominate at small $K$, while numerical approximations can produce nonzero bias $b_\eta$.



\textbf{Finite-Compute Estimator Correction.} Rather than modifying the Monte Carlo solver, we treat its finite-budget output as a noisy observation $\hat u_K=u+b_\eta+\xi_K$, where $\xi_K$ is zero-mean sampling noise with spatially varying variance and nearly zero measured spatial error autocorrelation. The corrected estimator is $\tilde u=\Phi_\theta(\hat u_K,f)$, evaluated for error reduction at fixed compute. For an idealized squared-loss population objective, $\Phi^\star(\hat u_K,f)=\mathbb E[u\mid\hat u_K,f]$ we train with:
\begin{equation}
\small
\min_\theta\;
\frac{\sum_{i,q}M_i(q)|\Phi_\theta(\hat U_{i,K},F_i)(q)-U_i(q)|}
{\sum_{i,q}M_i(q)+10^{-8}},
\end{equation}
using ground truth solutions as supervision. Here $U_i$, $(\hat U_{i,K},F_i)$, and $M_i$ are normalized targets, normalized input pairs, and interior masks where $i$ indexes instances in a training batch and $q$ indexes grid points. At a fixed PDE instance, MSE improves exactly when
\[
V_{\rm raw}-V_{\rm corr}>b_{\rm corr}^2-b_{\rm raw}^2,
\]
where $V$ and $b^2$ denote variance and squared bias, respectively. The model has exactly two channels $(\hat u_K,f)$, with masks used in preprocessing, loss, and evaluation. The learned correction is a post-processing step and does not separately ingest PDE coefficients, boundary values, or solver internals.

\textbf{Field Discretization.} For learning and evaluation, continuous solution fields are discretized on uniform grids, yielding arrays $U \in \mathbb{R}^{H \times W}$. Each training example consists of a low-budget Monte Carlo estimator $\hat U_K$, its source field $F$, and a reference solution $U^\star$. 

\textbf{Neural Estimator Correction.} We parameterize $\Phi_\theta$ by using a modified SUNet (Swin-UNet), a transformer-based encoder–decoder originally developed for image restoration. Its hierarchy, attention, and skip connections combine spatial solution structure with the heterogeneous noise level.
At \textbf{inference time,} we run a Monte Carlo solver at budget $K$ to obtain $\hat u_K$ and compute $\tilde u=\Phi_\theta(\hat u_K,f)$ in one forward pass.


\section{PDEZoo: A Large-Scale Benchmark for Elliptic PDEs}


\begin{table*}[t]
\centering
\footnotesize
\setlength{\tabcolsep}{3.5pt}
\renewcommand{\arraystretch}{1.08}
\caption{Comparison of PDE benchmarks. PDEZoo combines exact analytic ground truth, diverse per-instance geometries, multi-budget stochastic solver data, and a scalable procedural generator. Its scale denotes the expanded training corpus; the reported MC$^2$ experiments use the original 100,000-instance pool.}
\label{tab:benchmark_comparison}
\begin{tabularx}{\textwidth}{@{}
p{1.8cm}
>{\raggedright\arraybackslash}X
p{1.9cm}
c
c
c
r
@{}}
\toprule
Benchmark 
& PDE Scope 
& Geometries
& \makecell{Analytic\\[-1pt]Ground Truth}
& \makecell{Intermediate\\[-1pt]Solutions}
& \makecell{Procedural\\[-1pt]Generator}
& Scale \\
\midrule
PDEBench \cite{takamoto2022pdebench} 
& Time-dep.\ PDEs 
& Fixed grids
& \ding{55} 
& \ding{55} 
& \ding{55}
& ${\sim}10$k \\
PDEArena \cite{gupta2022towards}
& Time-dep.\ PDEs 
& Fixed grids
& \ding{55} 
& \ding{55} 
& \ding{55}
& ${\sim}5$--$10$k \\
FNO Data \cite{li2021fourier}
& Canonical PDEs 
& Unit square
& \ding{55} 
& \ding{55} 
& \ding{55}
& $\sim7$k \\
Poseidon \cite{herde2024poseidon}
& Multi-PDE 
& Fixed grids
& \ding{55} 
& \ding{55} 
& \ding{55}
& $1.6$M \\
\midrule
\textbf{PDEZoo} 
& Elliptic (3 training; 5 test families)
& \textbf{\makecell[l]{composite \\from primitives}}
& {\color{green!70!black}\ding{51}} 
& {\color{green!70!black}\ding{51}} 
& {\color{green!70!black}\ding{51}}
& \textbf{2M} (train) \\
\bottomrule
\end{tabularx}
\end{table*}



We introduce \textbf{PDEZoo}, a large-scale benchmark for elliptic partial differential equations (PDEs). Each instance provides (i) an explicit elliptic boundary-value problem with Dirichlet conditions for second-order equations and Navier conditions for biharmonic problems, (ii) a known \emph{analytic} solution used as ground truth, and (iii) optional multi-budget trajectories produced by a Monte Carlo solver.

PDEZoo combines exact analytic ground truth, diverse per-instance geometries, multi-budget stochastic solver data, and a scalable procedural generator (Table \ref{tab:benchmark_comparison}), allowing us to benchmark solver performance at various compute levels within a unified and reproducible evaluation protocol.

\subsection{Problem Specification}
\label{subsec:PDEZoo_problem}

Each PDE instance is specified by $\mathcal P=(\Omega,\mathcal E,\vartheta,f,\mathcal G,u)$, where $\Omega\subset\mathbb R^2$ is the domain, $\mathcal E$ the family, $\vartheta$ its scalar parameter when present, $f$ the source, $\mathcal G$ the complete boundary data, and $u$ the analytic reference.
PDEZoo includes five elliptic PDE families:
\[
\begin{alignedat}{6}
\text{Laplace:}\quad     & \Delta u = 0,
&\qquad
\text{Poisson:}\quad     & \Delta u = f,
&\qquad
\text{Yukawa:}\quad      & \Delta u - \lambda u = f, \\
\text{Biharmonic:}\quad  & \Delta^2 u = f,
&\qquad
\text{Helmholtz:}\quad   & \Delta u + k^2 u = f.
\end{alignedat}
\]
Second-order families use $u=g$ on $\partial\Omega$ and biharmonic uses the Navier pair $u=g_1$ and $\Delta u=g_2$ there.
All source and boundary data are derived exactly from $u$, including $(g_1,g_2)=(u,\Delta u)|_{\partial\Omega}$ for biharmonic.
Additional details on the manufactured solution families and parameter ranges are provided in Appendix~\ref{subsec:PDEZoo_problem_append}.

\subsection{PDEZoo Generation}
\label{subsec:pdezoo_overview}

First, we sample a PDE family (Laplace, Poisson, or Yukawa, each with probability $\nicefrac{1}{3}$) and a domain $\Omega$, with the latter drawn from eight primitive families, each with probability $1/11$, or a Boolean composition of two primitives with probability $3/11$. We then construct an analytic solution $u$ via the \emph{method of manufactured solutions}: $u$ is defined as a random sum of $n$ atom functions drawn from a curated pool, and the forcing term $f$ and boundary condition $g$ are derived analytically from $u$ and the selected PDE family. Test packs are generated separately for all five families.

\textbf{Solution atoms.}
The atom pool is designed to span the spectrum of features that arise in real elliptic PDE solutions: smooth low-frequency modes (polynomials, plane waves), localized structures (Gaussian bumps, radial basis functions), singularity-like behavior (log sources, rational functions), and sharp gradients (tanh transitions, high-frequency products). 

\textbf{Domains.}
Domains are represented implicitly through signed distance functions (SDFs), which support geometry evaluation, boundary projection, and interior/boundary sampling. PDEZoo includes eight primitive domain families (disk, square, rectangle, ellipse, annulus, triangle, hexagon, and stadium) as well as composed domains formed by Boolean operations (union, intersection, difference) on exactly two primitives. Composed SDFs are conservative underestimates of the true boundary distance, providing conservative walk radii (Appendix~\ref{app:geometry}).

\textbf{Ground truth.}
Ground-truth labels are given by evaluating the analytic solution $u$ on a fixed $256 \times 256$ grid over pixel space.

\subsection{Finite Compute Monte Carlo Partial Solutions}
\label{subsec:PDEZoo_mc_main}
For each PDE instance, PDEZoo includes intermediate Monte Carlo WoS solver outputs at a set of budgets
\begin{figure}
    \centering
    \includegraphics[width=1\linewidth]{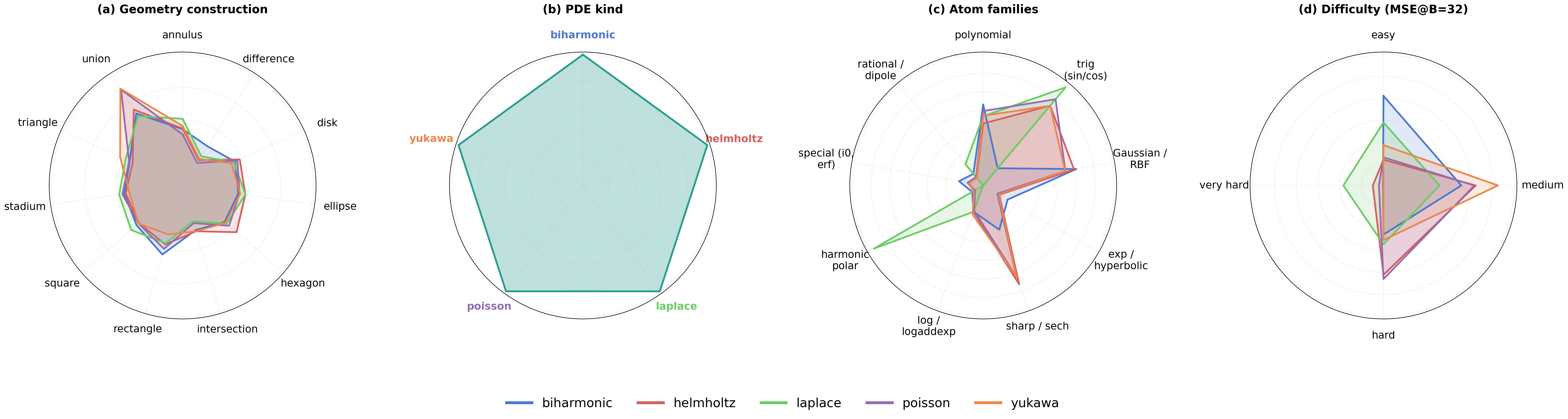}
    \caption{Overview of Data Distribution PDEZoo}
    \label{fig:pdezoo_overview}
\end{figure}
$\mathcal{B}=\{B_1<\cdots<B_m\}$, producing a trajectory $\{\hat{u}_{B}\}_{B\in\mathcal{B}}$ evaluated on the same grid as $u$.
These trajectories expose finite-sample estimator behavior and allow models to be trained or evaluated using intermediate solutions. Budget sets, storage format, and reproducible generation details are in Appendix \ref{app:format}.

\subsection{Splits, Tasks, and Evaluation Protocol}
\label{subsec:PDEZoo_eval_main}

\textbf{Fixed benchmark splits.}
PDEZoo uses fixed training and test splits for reproducible evaluation without training on test cases. The expanded training split contains 2,000,000 Laplace, Poisson, and Yukawa instances with WoS solutions at budgets $\mathcal{B}_\text{train}=\{1,2,4,8,16,32\}$. The reported MC$^2$ experiments use the original 100,000-instance pool, with 5,000 for validation. The test split contains 5,000 instances across Laplace, Poisson, Yukawa, Helmholtz, and Biharmonic, with WoS solutions at 17 budgets $B\in\{2^j:j=1,\ldots,17\}$. Helmholtz and Biharmonic instances are reserved for testing to evaluate generalization across PDE types. The stored data include $256\times256$ grids of the WoS estimate, analytic solution, and source, together with the domain mask. 

\textbf{Evaluation Settings.}
PDEZoo supports three evaluation settings on each PDE type:\\
(a) \text{Dense-field solving:} predict $u$ on the evaluation grid given $(\Omega,\mathcal{E},\vartheta,f,\mathcal G)$ \\
(b) \text{Finite-compute correction:} given a low-budget estimate $\hat{u}_{B}$, produce an improved estimate under fixed compute. \\
(c) \text{Point-query evaluation:} estimate $u(x)$ at queried points $x\in\Omega$. 


\textbf{Difficulty calibration.}
A useful benchmark requires a controlled spread of instance difficulty. We classify instances into four tiers based on MSE at the reference budget $B\!=\!32$ (Table~\ref{tab:difficulty_main}). See Appendix ~\ref{app:difficulty} for more details.

\section{Experiments}

\begin{figure}
    \centering
    \includegraphics[width=0.75\linewidth]{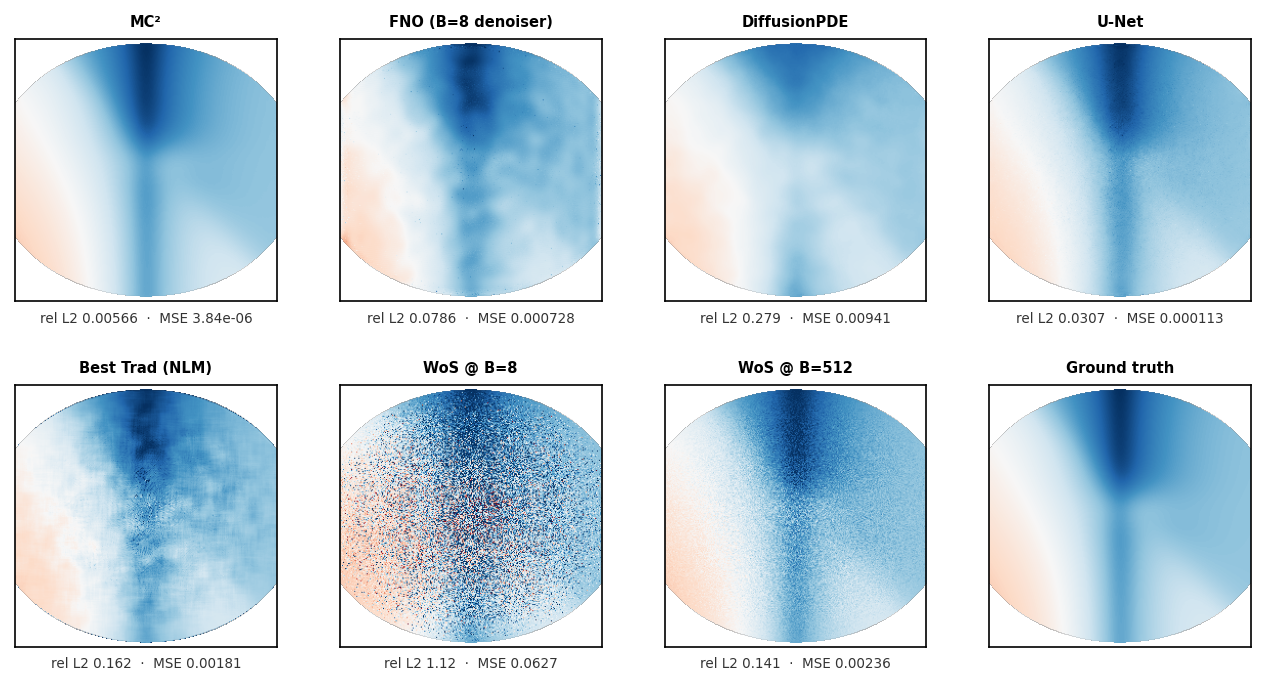}
    \caption{\textbf{Qualitative comparison of MC$^2$ against baselines on a 
    Poisson instance.} MC$^2$ closely matches analytic 
    ground truth and neural operators perform poorly. 
    Raw WoS at $B\!=\!8$ retains substantial noise, and the best traditional denoiser over-smooths boundaries. }
\label{fig:qualitative}
\end{figure}


\subsection{Training Regime}

We train all methods on the same random subset of 100,000 elliptic PDE from PDEZoo's training split. Each instance consists of a PDE specification (forcing function $f$, boundary condition $g$, and domain $\Omega$), a Monte Carlo Walk-on-Spheres (WoS) solution at budget of $B=8$, and the corresponding analytic ground truth solution $u$. We use a subset of 5,000 PDEs from the PDEZoo training split for training validation, and use it for early stopping and hyperparameter selection. 

For MC$^2$, we train using data from MC budgets of $B\in\{1,2,4,8,16,32\}$ and their respective source terms, with an error-correction objective using our analytical ground truth as the target output.  During inference, the $B=8$ WoS solutions (noisy estimates) and the source term are inputted into the model. 
 
For baselines trained to learn operators, the inputs are the forcing functions $f$ evaluated on the same grid. For diffusion baselines, we train a separate EDM teacher per PDE family on the joint distribution $(u,f,\lambda,m)$ with the standard score-matching loss. WoS observations are not used for teacher training, but both guided samplers receive the $B=8$ WoS field and boundary values at inference.

\subsection{Comparisons}
\label{subsec:baselines}

\textbf{Walk-on-Spheres (WoS).}
We evaluate stored WoS budgets $B=2^j$, $j=1,\ldots,17$.  Ideal finite-variance sampling has $O(B^{-1})$ variance, and we note that the implemented solver also has truncation errors. 

\textbf{Traditional Denoisers.}
We apply four classical denoising algorithms to budget $B=8$ WoS outputs: Gaussian filtering, heavy Gaussian filtering, total variation, and non-local means.  These methods assume generic noise models and do not leverage PDE-specific structure.

\textbf{Neural Operator Models.}
We evaluate FNO \cite{li2021fourier} and PINO \cite{DBLP:journals/corr/abs-2111-03794} in the operator learning setting, where the input is the forcing function $f$ and the target is the solution $u$. PINO uses the same FNO architecture with an additional physics loss term, $\mathcal{L} = \mathcal{L}_{\mathrm{data}} + \lambda \mathcal{L}_{\mathrm{physics}}$ with $\lambda = 0.1$,
where $\mathcal{L}_{\mathrm{data}}$ is masked relative-$L^2$ error and $\mathcal{L}_{\mathrm{physics}}$ is the implemented discrete PDE-residual penalty. We additionally evaluate a matched parameter U-net with inputs $(\hat u_8, f)$ and a FNO model with inputs $(f,g,m)$, and .

\textbf{Diffusion Models.} Our DiffusionPDE-mode and CoCoGen-mode evaluations share a per-family EDM teacher and use different guided samplers, both conditioned on $B=8$ WoS observations and boundary values. We report results on Laplace, Poisson, and Yukawa, with no checkpoints for the held-out families.



\paragraph{Evaluation protocol.}\label{subsec:eval_metrics}
We evaluate over valid domain pixels $V$. For each case, we normalize the ground truth and prediction using the ground-truth minimum and maximum, then clip the prediction to $[0,1]$. Denoting these normalized fields by $u_{01}$ and $p_{01}$, we compute
\[
\small
\mathrm{MSE}=\frac{1}{|V|}\sum_{i\in V}(p_{01,i}-u_{01,i})^2,\quad
\mathrm{PSNR}=-10\log_{10}(\mathrm{MSE}),\quad
\mathrm{SNR}=10\log_{10}\left(
\frac{\sum_{i\in V}u_{01,i}^2}{|V|\cdot\mathrm{MSE}}
\right).
\]
Raw MSE and relative $L^2$ error use the original fields without normalization or clipping. SSIM is averaged over valid pixels within the domain's bounding box. LPIPS uses AlexNet, with invalid pixels set to zero after normalization.

\subsection{Evaluation Results}
 
\begin{table}[t]
  \centering
  \footnotesize
  \setlength{\tabcolsep}{3pt}
  \caption{MC$^2$ matches the accuracy of substantially more WoS walks and outperforms the evaluated neural baselines on this benchmark. Evaluation uses the 5,000-instance PDEZoo test split.  Unannotated FNO/PINO are $f$-only operators and correctors receive B8 observations. Diffusion models do not generalize to unseen PDE families and are only evaluated on the average of Laplace, Yukawa, and Poisson. Metrics follow Section~\ref{subsec:eval_metrics}.  Arrows indicate best metric direction}
  \begin{sc}
    \begin{tabular}{@{}lcccc@{}}
      \toprule
      Method & PSNR (dB) $\uparrow$ & SNR (dB) $\uparrow$ & MSE $\downarrow$ & LPIPS $\downarrow$ \\
      \midrule
      \multicolumn{5}{l}{\textit{Walk-on-Spheres (Monte Carlo)}} \\
      WoS ($B\!=\!8$)        & $15.26 \pm 7.43$ & $9.773 \pm 7.527$ & $(8.02 \pm 8.34) \times 10^{-2}$ & $0.7275 \pm 0.4198$ \\
      WoS ($B\!=\!16384$)    & $42.14 \pm 13.82$ & $36.644 \pm 13.843$ & $(3.13 \pm 12.43) \times 10^{-3}$ & $0.1964 \pm 0.3088$ \\
      WoS ($B\!=\!32768$)    & $44.98 \pm 13.92$ & $39.486 \pm 13.935$ & $(1.88 \pm 8.77) \times 10^{-3}$ & $0.1592 \pm 0.2731$ \\
      WoS ($B\!=\!65536$)    & $47.80 \pm 14.00$ & $42.307 \pm 14.009$ & $(1.11 \pm 6.15) \times 10^{-3}$ & $0.1263 \pm 0.2381$ \\
      \midrule
      \multicolumn{5}{l}{\textit{Traditional Denoisers (applied to $B\!=\!8$ WoS)}} \\
      Gaussian               & $18.85 \pm 8.31$ & $13.361 \pm 8.253$ & $(5.06 \pm 6.96) \times 10^{-2}$ & $0.4032 \pm 0.2520$ \\
      Gaussian (heavy)       & $17.66 \pm 6.48$ & $12.164 \pm 6.307$ & $(4.06 \pm 5.04) \times 10^{-2}$ & $0.2832 \pm 0.2062$ \\
      Total Variation        & $24.63 \pm 10.41$ & $19.141 \pm 10.419$ & $(2.71 \pm 4.86) \times 10^{-2}$ & $0.2613 \pm 0.2123$ \\
      Non-Local Means        & $26.63 \pm 10.36$ & $21.134 \pm 10.444$ & $(1.93 \pm 3.94) \times 10^{-2}$ & $0.3212 \pm 0.3014$ \\
      \midrule
      \multicolumn{5}{l}{\textit{Operator-learning baselines, no MC observation (input $f$ only)}} \\
      FNO                    & $7.74 \pm 5.47$ & $2.251 \pm 4.846$ & $(2.58 \pm 1.57) \times 10^{-1}$ & $0.3790 \pm 0.2277$ \\
      PINO                   & $8.24 \pm 6.13$ & $2.745 \pm 5.878$ & $(2.49 \pm 1.57) \times 10^{-1}$ & $0.3753 \pm 0.2236$ \\
    \multicolumn{5}{l}{\textit{Operator-learning baselines, no MC observation (input $(f,g,m)$ only)}} \\
  
\makecell[l]{FNO operator\\$(f,g,m)$} & $18.04 \pm 11.28$ & $12.55 \pm 11.25$ & $(7.98\pm10.80)\times10^{-2}$ & $0.2243 \pm 0.1977$ \\
\makecell[l]{PINO operator\\$(f,g,m)$} & $7.95 \pm 5.28$ & $2.46 \pm 5.37$ & $(2.43\pm1.53)\times10^{-1}$ & $0.3863 \pm 0.2044$ \\
    \multicolumn{5}{l}{\textit{Learned correctors of the B = 8 WoS field (MC observation as input)}} \\
    
\makecell[l]{FNO corrector\\$(\hat u_8)$} & $31.00 \pm 11.72$ & $25.51 \pm 11.80$ & $(1.32\pm3.42)\times10^{-2}$ & $0.2002 \pm 0.2800$ \\
\makecell[l]{U-Net corrector\\$(\hat u_8)$} & $30.93 \pm 11.69$ & $25.44 \pm 11.77$ & $(1.33\pm3.50)\times10^{-2}$ & $0.1982 \pm 0.2764$ \\

\makecell[l]{FNO corrector\\$(\hat u_8,f)$} & $31.25 \pm 11.57$ & $25.76 \pm 11.65$ & $(1.27\pm3.50)\times10^{-2}$ & $0.1676 \pm 0.2560$ \\
\makecell[l]{U-Net corrector\\$(\hat u_8,f)$} & $37.32 \pm 12.92$ & $31.83 \pm 12.97$ & $(6.04\pm20.80)\times10^{-2}$ & $0.1997 \pm 0.2913$ \\
    \multicolumn{5}{l}{\textit{Diffusion baselines with $B=8$ WoS observations and boundary values}} \\
CoCoGen & $30.07 \pm 10.16$ & $24.476 \pm 10.300$ & $(8.80 \pm 24.51)\times10^{-3}$ & $0.0801 \pm 0.1236$ \\
DiffusionPDE & $30.42 \pm 9.99$ & $24.835 \pm 10.127$ & $(7.46 \pm 20.65)\times10^{-3}$ & $0.0967 \pm 0.1592$ \\

     \midrule
      \multicolumn{5}{l}{\textit{MC$^2$  input ablation with Budget = 8 only}} \\
      MC$^2$  ($\hat u_8, f, g, m$ only)       & $43.22 \pm 12.27$ & $37.72 \pm 12.33$ & $(2.26 \pm 12.38) \times 10^{-3}$ & $0.0410 \pm 0.1156$ \\
      MC$^2$  ($\hat u_8$ only)  & $40.04 \pm 13.58$ & $34.55 \pm 13.60$ & $(2.93 \pm 11.82) \times 10^{-3}$ & $0.0825 \pm 0.1909$ \\
      MC$^2$  ($f$ only)         & $10.65 \pm 5.76$ & $5.16 \pm 6.07$ & $(1.59 \pm 1.45) \times 10^{-1}$ & $0.3731 \pm 0.2268$ \\
      MC$^2$  ($f, g,m$ only)       & $9.58 \pm 4.85$ & $4.09 \pm 5.31$ & $(1.72 \pm 1.37) \times 10^{-1}$ & $0.3721 \pm 0.2129$ \\
      \midrule
      \textbf{MC$^2$ (Ours, Multibudget)} & $\mathbf{44.90 \pm 12.49}$ & $\mathbf{39.410 \pm 12.532}$ & $\mathbf{(1.94 \pm 11.38) \times 10^{-3}}$ & $\mathbf{0.0164 \pm 0.0681}$ \\
      \bottomrule
    \end{tabular}
  \end{sc}
  \label{tab:main_results}
\end{table}

We evaluate on the 5,000-instance PDEZoo test split and present our quantitative results across all methods in Table~\ref{tab:main_results}.  MC$^2$ reaches $44.90$ dB from $B=8$ inputs at $15.26$ dB, a $29.64$ dB gain. On the other hand, WoS at $B=32{,}768$ reaches $44.98$ dB with $4,096$ times as many walks per point. A parameter-matched U-Net reaches $37.32$ dB and FNO correctors reach $31.25$ dB. On the three in-distribution families, diffusion baselines conditioned on $B=8$ WoS observations and boundary values reach $30.07$ dB for CoCoGen-mode and $30.42$ dB for DiffusionPDE-mode, compared with $16.82$ dB for the WoS input and $49.55$ dB for MC$^2$ (Table~\ref{tab:per_family_psnr}). Though our main comparison focuses on mesh-free methods, we recognize the usefulness of comparison to mesh-based methods. Since our easily meshed 2D domains favor FEM without testing the benefits of avoiding volumetric meshing, Appendix~\ref{app:fem_results} compares both 2D and 3D settings, showing  where MC$^2$ offers advantages in runtime and robustness to meshing failures consistent with the weaknesses typical of mesh-based solvers.                   
  
\textbf{Generalization to Out-of-Distribution (OOD) PDE Families: }The same weights achieve $35.96$ dB on biharmonic and $40.05$ dB on Helmholtz (Table~\ref{tab:per_family_psnr}). This split shares manufactured-solution atoms across training/testing. Independent-reference and disjoint-atom tests appear in Appendices~\ref{app:fem_results} and~\ref{app:unseen_atoms}. 

\begin{table}[t]
  \centering
  \footnotesize
\caption{Per-family PSNR (dB, $\uparrow$) on the PDEZoo test set. MC$^2$ achieves the highest PSNR on every PDE family compared to other Neural Network methods, including the held-out Biharmonic and Helmholtz families absent from the training distribution. Diffusion rows use $B=8$ WoS observations and boundary values on the in-distribution cases, with no held-out-family checkpoint. Average columns pool cases within ID/OOD groups, rather than equally weighting family means.  MC$^2$ uses a single $B\!=\!8$ WoS input. Bold marks the best result per column among learned and post-processing methods. Full per-family breakdown on SNR, MSE and LPIPS can be found in Appendix \ref{app:full_per_family}.}
  \begin{sc}
  \label{tab:per_family_psnr}
  \setlength{\tabcolsep}{3pt}
  \adjustbox{max width=\linewidth}{%
    \begin{tabular}{@{}lccccccc@{}}
      \toprule
      & \multicolumn{4}{c}{\textit{In-distribution}} & \multicolumn{3}{c}{\textit{Out-of-distribution}} \\
      \cmidrule(lr){2-5} \cmidrule(lr){6-8}
      Method & Laplace & Poisson & Yukawa & Average & Biharmonic & Helmholtz & Average \\
      \midrule
      \multicolumn{8}{l}{\textit{Walk-on-Spheres (Monte Carlo)}} \\
      WoS ($B\!=\!8$)        & $25.22 \pm 3.33$  & $11.20 \pm 5.25$  & $13.89 \pm 5.88$  & $16.82 \pm 7.84$  & $15.01 \pm 6.38$  & $10.93 \pm 5.01$  & $12.96 \pm 6.08$ \\
      WoS ($B\!=\!8192$)     & $55.29 \pm 3.42$  & $32.94 \pm 11.73$ & $38.83 \pm 10.08$ & $42.41 \pm 13.18$ & $37.30 \pm 13.76$ & $32.05 \pm 11.88$ & $34.67 \pm 13.12$ \\
      WoS ($B\!=\!16384$)    & $58.30 \pm 3.42$  & $35.86 \pm 11.85$ & $41.81 \pm 10.11$ & $45.38 \pm 13.25$ & $39.77 \pm 13.91$ & $34.89 \pm 12.08$ & $37.32 \pm 13.25$ \\
      WoS ($B\!=\!32768$)    & $61.31 \pm 3.42$  & $38.81 \pm 11.92$ & $44.80 \pm 10.12$ & $48.37 \pm 13.29$ & $42.15 \pm 13.98$ & $37.77 \pm 12.22$ & $39.95 \pm 13.30$ \\
      \midrule
      \multicolumn{8}{l}{\textit{Traditional Denoisers (applied to $B\!=\!8$ WoS)}} \\
      Gaussian               & $29.67 \pm 2.56$  & $14.64 \pm 6.57$  & $17.57 \pm 6.09$  & $20.68 \pm 8.46$  & $18.11 \pm 7.59$  & $14.21 \pm 6.40$  & $16.15 \pm 7.28$ \\
      Gaussian (heavy)       & $25.98 \pm 3.30$  & $14.64 \pm 4.95$  & $16.55 \pm 4.87$  & $19.10 \pm 6.67$  & $16.73 \pm 5.82$  & $14.32 \pm 4.98$  & $15.52 \pm 5.54$ \\
      Total Variation        & $37.81 \pm 2.63$  & $19.34 \pm 8.60$  & $23.34 \pm 7.77$  & $26.89 \pm 10.51$ & $23.81 \pm 9.62$  & $18.80 \pm 8.29$  & $21.29 \pm 9.32$ \\
      Non-Local Means        & $39.16 \pm 3.05$  & $21.56 \pm 8.85$  & $25.54 \pm 7.71$  & $28.81 \pm 10.31$ & $25.91 \pm 9.90$  & $20.88 \pm 8.48$  & $23.39 \pm 9.55$ \\
      \midrule
      \multicolumn{8}{l}{\textit{Neural Network Methods}} \\
      FNO                    & $16.52 \pm 5.94$  & $5.47 \pm 1.84$   & $5.68 \pm 1.93$   & $9.27 \pm 6.41$   & $5.43 \pm 2.36$   & $5.52 \pm 1.99$   & $5.47 \pm 2.18$ \\
      PINO                   & $18.46 \pm 6.15$  & $5.71 \pm 1.95$   & $5.65 \pm 1.89$   & $10.00 \pm 7.20$  & $5.54 \pm 2.32$   & $5.70 \pm 2.19$   & $5.62 \pm 2.26$ \\
FNO $(f,g,m)$ & $36.79 \pm 6.11$ & $14.77 \pm 5.11$ & $14.84 \pm 5.17$ & $22.25 \pm 11.76$ & $8.94 \pm 7.10$ & $14.68 \pm 5.01$ & $11.82 \pm 6.78$ \\
PINO $(f,g,m)$ & $16.31 \pm 5.47$ & $6.04 \pm 2.09$ & $5.97 \pm 2.01$ & $9.49 \pm 6.05$ & $5.25 \pm 2.57$ & $6.08 \pm 2.32$ & $5.67 \pm 2.48$ \\
FNO corrector $(\hat u_8,f)$ & $44.12 \pm 5.04$ & $26.86 \pm 9.98$ & $30.97 \pm 8.44$ & $34.03 \pm 10.96$ & $28.41 \pm 12.17$ & $25.85 \pm 10.05$ & $27.12 \pm 11.23$ \\
U-Net corrector $(\hat u_8,f)$ & $53.19 \pm 2.80$ & $32.11 \pm 10.74$ & $36.93 \pm 9.26$ & $40.81 \pm 12.31$ & $33.30 \pm 12.97$ & $31.02 \pm 10.95$ & $32.16 \pm 12.05$ \\

      CoCoGen & $33.31 \pm 12.28$ & $27.11 \pm 8.76$ & $29.76 \pm 7.80$ & $30.07 \pm 10.16$ & -- & -- & -- \\
DiffusionPDE & $34.40 \pm 11.56$ & $27.12 \pm 8.77$ & $29.70 \pm 7.75$ & $30.42 \pm 9.99$ & -- & -- & -- \\
      \midrule
      \textbf{MC$^2$ (Ours)} & $\mathbf{59.94 \pm 3.85}$ & $\mathbf{42.17 \pm 9.54}$ & $\mathbf{46.40 \pm 8.21}$ & $\mathbf{49.55 \pm 10.75}$ & $\mathbf{35.96 \pm 13.25}$ & $\mathbf{40.05 \pm 9.42}$ & $\mathbf{38.01 \pm 11.67}$ \\
      \bottomrule
    \end{tabular}}
  \end{sc}
\end{table}




\subsection{MC$^2$ Ablations}
\label{sec:experiments}

\textbf{Multi-budget vs.\ single-budget training.}
Our MC$^2$ model is trained jointly on noisy Walk-on-Spheres solutions at budgets $B \in \{1,2,4,8,16,32\}$. On the same $B=8$ test inputs, multi-budget and full-data single-budget models achieve $44.90$ and $44.96$ dB. 

\textbf{Source-term conditioning.}
To quantify the importance of conditioning on the PDE source term $f$, we compare models trained without $f$ against the full MC$^2$ model at the same budget. The controlled $B=8$ ablation gives median PSNR $44.29$ dB with $f$ and $40.56$ dB without $f$ while retaining the mask.  


\textbf{Loss precision.}
We compare FP32 and BF16 no-source training ablations at B8.  The FP32-loss model reaches $57.7$ dB PSNR validation score, compared to $54.8$ dB validation score for the BF16-loss model. 


\textbf{Data scaling.}
We train five MC$^2$ models using $4\%$, $8\%$, $16\%$, $32\%$, and $64\%$ of the training set. All runs use the same hyperparameters. These runs trace a data-scaling curve for MC$^2$ at a fixed Monte Carlo input budget.
\begin{figure}
    \centering
    \includegraphics[width=1\linewidth]{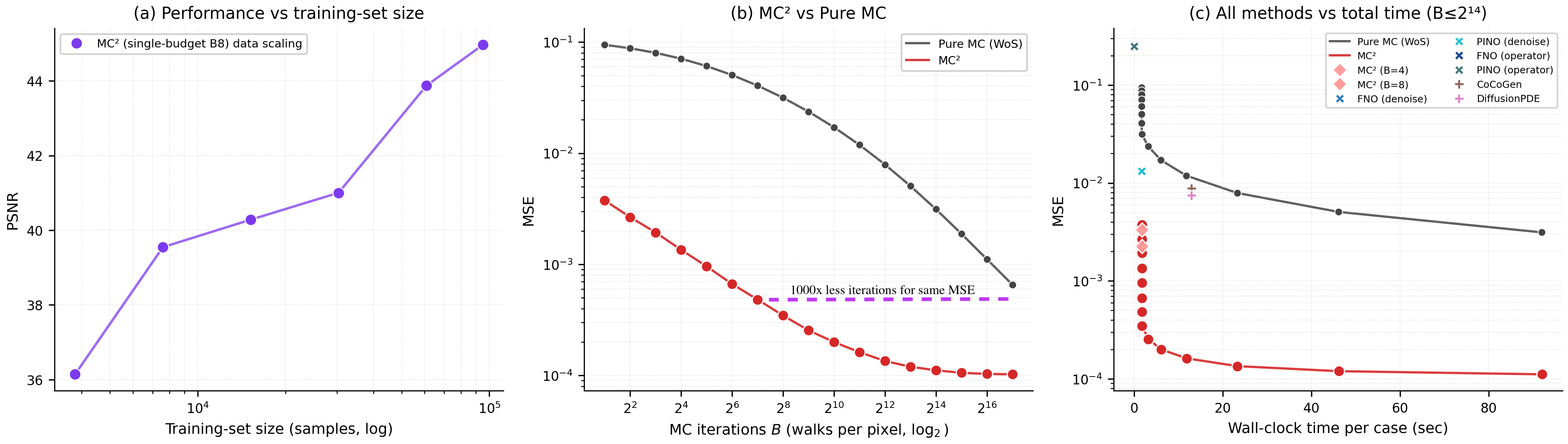}
    \caption{\textbf{Scaling Study of MC$^2$} (a) Single-budget B8 performance improves with more training examples. (b) We evaluate performance of applying MC$^2$ error corrector to different WoS budgets. The aggregate corrected error is smaller over the measured range but approaches a nonzero floor. (c) Accuracy--compute tradeoff. }
\label{fig:scaling}
\end{figure}

\section{Discussion}

\textbf{Finite-sample Monte Carlo error is structured, learnable, and correctable.}
The measured WoS errors have nearly zero spatial autocorrelation but strongly spatially varying variance. The following comparisons characterize the roles of stochastic observations, source information, and learned solution priors without isolating them completely:

\begin{itemize}[leftmargin=*,itemsep=2pt,topsep=2pt]
\item \textbf{Observation dependence.} Controlled same-architecture ablations give median PSNR 44.29 dB with $(\hat u_8,f,m)$, 40.56 dB with $(\hat u_8,m)$, and 8.56 dB with $(f,g,m)$. The corrector therefore exploits joint statistics of $(\hat{u}_K, u)$ that are not recoverable from the PDE specification alone. 
\item \textbf{Operator transfer.} The model achieves 35.96 and 40.05 dB on held-out biharmonic and Helmholtz families, consistent with learning shared features of finite-sample WoS error, such as geometry-dependent variance, though shared solution atoms prevent ruling out learned solution priors.    
\end{itemize}

We emphasize the scope of this claim. Our evidence establishes that finite-sample WoS error on 2D elliptic problems contains learnable structure sufficient to recover $\sim$30~dB from a single $B\!=\!8$ estimate. However, we do not claim the structure admits a closed-form description, that the learned corrector is Bayes-optimal, or that comparable gains transfer to arbitrary stochastic solvers. Appendix~\ref{app:bias_variance} states the bias--variance condition for MSE improvement and reports the measured tradeoff.

\textbf{Beyond Walk-on-Spheres: Potential Extensions.}
Although we instantiate Monte Carlo error correction with WoS, nothing in the framework is specific to PDEs or to walk-based estimators. The key requirements are simply (i) a stochastic estimator $\hat{u}_K$ whose finite-budget error has learnable structure and (ii) access to high-fidelity reference outputs or ground truths for supervision. Potential application areas include path-traced rendering, Markov chain Monte Carlo posterior sampling, particle filters, neutron transport simulations, and quantum Monte Carlo. Learning to post-process Monte Carlo estimates has precedents in rendering \citep{rousselle2013robust,bako2017kpcn,chaitanya2017recurrent,gharbi2019samplebased}. Our contribution concerns source-conditioned correction of finite-budget WoS fields and extending these to other domains is immediate future work. 



\section{Conclusion}

We make two contributions to elliptic PDE solving in the finite-compute regime. \textbf{MC$^2$} establishes that finite-sample Monte Carlo error is structured and correctable in a single forward pass, achieving the accuracy of 1000x as many WoS walks at comparable mean PSNR and outperforming the evaluated learned and classical denoising baselines, including on out-of-distribution PDE families. \textbf{PDEZoo} provides multi-budget evaluation infrastructure and an expanded 2M-instance training corpus, separate from the original pool used for the headline model with analytic ground truth and multi-budget Monte Carlo trajectories. A main limitation is that we can only solve the class of PDEs that Walk-on-Spheres can solve, which are broadly elliptic problems on domains where the Green's-function machinery applies, and extending the framework to hyperbolic, parabolic, or strongly nonlinear regimes will require pairing it with a suitable numerical estimator. Within that scope, however, MC$^2$ closes most of the gap to high-budget Monte Carlo at a fraction of the cost, and we see the same recipe as a template for the broader family of unbiased Monte Carlo solvers.


\bibliographystyle{plainnat}
\bibliography{references}

\newpage 
\appendix

\section{PDEZoo: Detailed Specifications}

\subsection{Solution Atom Types}
\label{app:atoms}

Each solution $u(x,y)$ is constructed as a sum of $n$ randomly chosen atoms, where $n$ is sampled uniformly from $[2, 6]$. For Poisson and Yukawa instances, atoms are drawn from a \textbf{general pool} of 20 types. For Laplace instances, atoms are drawn from a \textbf{harmonic pool} of 12 unique types, real and imaginary parts of holomorphic functions such as $z^n$, $e^{kz}$, $\log(z-z_0)$, and $1/(z-z_0)^n$, guaranteeing $\nabla^2 h = 0$ by construction. High-frequency and sharp-featured atoms are deliberately over-represented to produce instances where Monte Carlo variance is large and correction is non-trivial, yielding a calibrated difficulty distribution (Appendix~\ref{app:atoms}). Combinatorially, choosing 2--6 atoms from these pools yields over 60,000 distinct type tuples for Poisson/Yukawa alone. Each tuple is further parameterized by continuous amplitudes, frequencies, and source locations, making the effective solution space inexhaustible.

\textbf{General atoms (Poisson/Yukawa).}
Table~\ref{tab:general_atoms} lists all 20 general atom types. Five of these are designated as \emph{hard atoms} (marked with $\dagger$), producing high-frequency oscillations, narrow features, or sharp transitions that increase Monte Carlo variance. Hard atoms are over-represented in the selection pool via the \texttt{hard\_atom\_extra} parameter (set to 4), which appends 4 extra copies of each hard atom, raising their draw probability from 25\% to 63\%.

\begin{table}[h]
\centering
\footnotesize
\caption{General atom types for Poisson and Yukawa instances. $\dagger$ = hard atom.}
\label{tab:general_atoms}
\begin{tabular}{@{}llp{5.5cm}@{}}
\toprule
Atom & Form & Key Parameters \\
\midrule
\texttt{poly} & Polynomial (degree $\leq 3$) & Coefficients in $[-1,1]$ \\
\texttt{trig} & $a \sin(k\pi x)$ & $k \in [1,6]$ \\
\texttt{exp} & $a \exp(bx)$ & $b \in [-2,2]$ \\
\texttt{log} & $a \log(x^2+0.01)$ & --- \\
\texttt{hyper} & $a \sinh/\cosh/\tanh(x)$ & --- \\
\texttt{special} & $a \operatorname{erf}(x)$ or $a\sqrt{x^2+\varepsilon}$ & --- \\
\texttt{harmonic\_polar} & $a r^n \cos(n\theta)$ & $n \in [1,6]$ \\
\texttt{plane\_wave} & $a \cos(k \hat{n}\cdot x)$ & $k \in [1,8]$ \\
\texttt{yukawa\_i0} & $a I_0(\mu r)$ & $\mu$ from $\lambda \in [0.5,6]$ \\
\texttt{log\_source} & $a \log|x - x_0|$ & $\rho \in [1.8,2.5]$ \\
\texttt{gaussian\_bump} & $a \exp(-w\|x-x_0\|^2)$ & $w \in [0.8,2]$ \\
\texttt{rational} & $a / (1+bx^2+cy^2)$ & $b,c \in [0.5,3]$ \\
\texttt{product} & $a f_1(k_1 x) f_2(k_2 y)$ & $f_i \in \{\sin,\cos,\tanh\}$ \\
\texttt{multi\_rbf} & Sum of 2--4 Gaussians & $w \in [2,8]$ \\
\texttt{high\_freq\_trig}$^\dagger$ & $a \sin(k_1\pi x)\cos(k_2\pi y)$ & $k_1,k_2 \in [2,8]$ \\
\texttt{logsumexp} & $a \operatorname{logaddexp}(bx, cy)$ & $b,c \in [1,4]$ \\
\texttt{sech\_bump}$^\dagger$ & $a / (\cosh^2 b(x\!-\!x_0) \cosh^2 c(y\!-\!y_0))$ & $b,c \in [2,6]$ \\
\texttt{very\_high\_freq\_trig}$^\dagger$ & $a \sin(k_1\pi x)\cos(k_2\pi y)$ & $k_1,k_2 \in [6,15]$ \\
\texttt{narrow\_bump}$^\dagger$ & $a \exp(-w\|x-x_0\|^2)$ & $w \in [8,25]$ \\
\texttt{sharp\_transition}$^\dagger$ & $a \tanh(k(\hat{n}\cdot x - c))$ & $k \in [5,15]$ \\
\bottomrule
\end{tabular}
\end{table}

\textbf{Harmonic atoms (Laplace).}
Table~\ref{tab:harmonic_atoms} lists the 12 unique harmonic atom types. Every harmonic atom is the real or imaginary part of a holomorphic function $f(z)$ where $z = x + iy$, guaranteeing $\nabla^2 h = 0$ analytically by the Cauchy--Riemann equations. Two types (marked $\dagger$) constitute the hard harmonic atoms, producing solutions with high spatial frequency or signal concentrated near the boundary.

\begin{table}[h]
\centering
\footnotesize
\caption{Harmonic atom types for Laplace instances. $\dagger$ = hard atom. Weight indicates the number of entries in the base selection pool.}
\label{tab:harmonic_atoms}
\begin{tabular}{@{}lllc@{}}
\toprule
Atom & Form & Complex Origin & Wt. \\
\midrule
\texttt{h\_polar} & $a r^n \cos/\sin(n\theta)$ & $\operatorname{Re/Im}(z^n)$, $n \in [1,6]$ & 2 \\
\texttt{h\_exp\_trig} & $a e^{kx} \cos/\sin(ky)$ & $\operatorname{Re/Im}(e^{kz})$, $k \in [-3,3]$ & 2 \\
\texttt{h\_trig\_hyp} & $a \sin(kx)\sinh(ky)$ etc. & Separable, $k \in [0.5,4]$ & 2 \\
\texttt{h\_log\_source} & $a \log|x - x_0|$ & $\operatorname{Re}(\log(z-z_0))$, $\rho \geq 1.8$ & 2 \\
\texttt{h\_linear} & $ax + by$ & $\operatorname{Re}(\alpha z)$ & 1 \\
\texttt{h\_bilinear} & $axy$ & $\operatorname{Im}(z^2)/2$ & 1 \\
\texttt{h\_arctan} & $a \operatorname{atan2}(y\!-\!y_0, x\!-\!x_0)$ & $\operatorname{Im}(\log(z\!-\!z_0))$ & 1 \\
\texttt{h\_inversion} & $a (x\!-\!x_0)/|x\!-\!x_0|^2$ & $\operatorname{Re}(1/(z\!-\!\bar{z}_0))$ & 1 \\
\texttt{h\_dipole} & $a (\Delta x^2\!-\!\Delta y^2)/|\Delta|^4$ & $\operatorname{Re}(1/(z\!-\!z_0)^2)$ & 1 \\
\texttt{h\_quadratic} & $a(u^2\!-\!v^2)$ or $2uv$ & $\operatorname{Re/Im}(z^2)$ rotated & 2 \\
\texttt{h\_high\_freq\_exp\_trig}$^\dagger$ & $a e^{kx} \cos/\sin(ky)$ & $\operatorname{Re/Im}(e^{kz})$, $k \in [4,10]$ & 2 \\
\texttt{h\_high\_n\_polar}$^\dagger$ & $a r^n \cos/\sin(n\theta)$ & $\operatorname{Re/Im}(z^n)$, $n \in [6,12]$ & 2 \\
\bottomrule
\end{tabular}
\end{table}

\subsection{Domain Geometry}
\label{app:geometry}

PDEZoo includes 8 primitive domain families and composed domains. All domains are centered approximately at the origin; the evaluation grid spans $[-1,1]^2$. Each primitive has probability $\nicefrac{1}{11}$ of being selected; composed domains have combined probability $\nicefrac{3}{11}$.

\begin{table}[h]
\centering
\footnotesize
\caption{Primitive domain types with parameter ranges and SDF properties.}
\label{tab:domains}
\begin{tabular}{@{}llll@{}}
\toprule
Domain & Key Parameters & SDF & Exact? \\
\midrule
Disk & $R = 1.0$ & $R - |x|$ & Yes \\
Square & half $= 1.0$ & $\min(\text{half} - |x_i|)$ & Yes \\
Rectangle & $a, b \in [0.6, 1.2]$ & $\min(a-|x|, b-|y|)$ & Yes \\
Ellipse & $r_x, r_y \in [0.6, 1.2]$ & $(1 - |x/r|)\min(r_x,r_y)$ & Approx. \\
Annulus & $r_\text{in} \in [0.2,0.6]$, $r_\text{out} \in [0.8,1.4]$ & $\min(r_\text{out}-r, r-r_\text{in})$ & Yes \\
Triangle & Equilateral, $R \in [0.6, 1.0]$ & $\min(\text{signed edge dists})$ & Yes \\
Hexagon & Regular, $R \in [0.6, 1.0]$ & $\min(\text{signed edge dists})$ & Yes \\
Stadium & half\_len $\in [0.2,0.6]$, rad $\in [0.25,0.5]$ & rad $-$ dist to segment & Yes \\
\bottomrule
\end{tabular}
\end{table}

\textbf{Composed domains.}
Composed domains combine two randomly selected primitives (from disk, rectangle, ellipse, or triangle) via a Boolean operation: union ($\nicefrac{3}{7}$), intersection ($\nicefrac{2}{7}$), or difference ($\nicefrac{2}{7}$). Each component has independent random parameters, rotation, and offset (separation $\in [0.15, 0.55]$). The composed SDF is computed as $\max(\text{sdf}_A, \text{sdf}_B)$ for union, $\min(\text{sdf}_A, \text{sdf}_B)$ for intersection, and $\min(\text{sdf}_A, -\text{sdf}_B)$ for difference. These are provably conservative (i.e., $\text{sdf}(x) \leq \operatorname{dist}(x, \partial\Omega)$), ensuring every WoS ball lies entirely within $\Omega$.



\subsection{Quality Filters}
\label{app:filters}

Each generated instance must pass eight quality filters before inclusion (up to 40 retries per case):

\begin{table}[h]
\centering
\footnotesize
\caption{Quality filters applied during case generation.}
\label{tab:filters}
\begin{tabular}{@{}lp{8cm}@{}}
\toprule
Filter & Condition \\
\midrule
Finite test point & $u(0.3, 0.4)$ is finite \\
Finite forcing & $f(0.3, 0.4)$ is finite (Poisson/Yukawa only) \\
Domain interior & $\geq 10$ interior points in 128 random samples \\
Composed thickness & Median SDF $> 0.03$ over interior points \\
Solution amplitude & $10^{-3} < \operatorname{std}(u) < 200$ \\
Boundary values & $\max|u(\partial\Omega)| < 500$ (composed only) \\
Forcing amplitude & $10^{-4} < \operatorname{std}(f) < 500$ (Poisson/Yukawa only) \\
Finite ground truth & All GT values inside domain are finite at $256 \times 256$ \\
\bottomrule
\end{tabular}
\end{table}

The WoS solver uses $\varepsilon=10^{-4}$. Shell, cap, and interpolation bias are treated jointly. For each instance, PDEZoo also provides an executable description of $u$ (and hence $f, g$) and geometry accessors (SDF evaluator, boundary projection, samplers).

\subsection{Data Format}
\label{app:format}

Each instance is stored as a JSONL record containing: case ID, PDE kind (\texttt{laplace}, \texttt{poisson}, \texttt{yukawa}, \texttt{helmholtz}, or \texttt{biharmonic}), operator parameter (the stored \texttt{lam} field denotes Yukawa $\lambda$ or Helmholtz $k^2$), a PyTorch expression string for the analytic solution, domain name and parameters, number of atom terms, and hardness metadata ($\operatorname{std}(u)$, $\operatorname{std}(f)$).

Each legacy \texttt{.npz} file contains three $(256 \times 256)$ float32 arrays:
\begin{itemize}
    \item \texttt{noisy}: MC solution at the given budget (0.0 outside domain)
    \item \texttt{clean}: Ground truth analytic solution (0.0 outside domain)
    \item \texttt{mask}: Domain indicator (1.0 inside, 0.0 outside)
\end{itemize}
The evaluated pipeline instead stores per-case PyTorch tensors under \texttt{gt/}, \texttt{masks/}, \texttt{solutions/}, \texttt{sources/}, and \texttt{bc/}, with solver settings and timing in \texttt{meta/}.

\subsection{Convergence Verification}
\label{app:convergence}

We fit log--log MSE slopes over budgets 1--32 on the baseline-generator diagnostic cases (167 per family with $\texttt{hard\_atom\_extra}=0$). Table~\ref{tab:convergence} reports the results: the median slope is exactly $-1.000$ for all three families, and 99.6\% of cases fall within $[-1.05, -0.95]$. This supports variance-dominated behavior over the tested range. Convergence verification results are presented in Table~\ref{tab:convergence} in the main text.

\subsection{Difficulty Distribution}
\label{app:difficulty}

Difficulty tier definitions and the observed distribution are presented in Table~\ref{tab:difficulty_main}. The difficulty distribution was calibrated via a 12-configuration hyperparameter sweep varying hard-atom weighting, term count ranges, and amplitude filter caps. The selected configuration (\texttt{hard\_atom\_extra}$\,=4$) shifts the distribution from a baseline of $(34/45/17/4)\%$ to the target $(20/41/30/8)\%$, ensuring sufficient representation of challenging instances.

\begin{table}[t]
\centering
\footnotesize
\caption{Difficulty tier definitions and observed distribution ($B\!=\!32$).}
\label{tab:difficulty_main}
\begin{tabular}{@{}lcl c@{}}
\toprule
Tier & MSE @ $B\!=\!32$ & Typical Characteristics & Proportion \\
\midrule
Easy & $< 10^{-2}$ & Smooth, low-frequency, simple domains & $\sim$20\% \\
Medium & $[10^{-2}, 1)$ & Moderate frequency, diverse domains & $\sim$41\% \\
Hard & $[1, 10^{2})$ & High-freq.\ oscillations, strong sources & $\sim$30\% \\
Very Hard & $\geq 10^{2}$ & Very high-freq.\ patterns, sharp features & $\sim$8\% \\
\bottomrule
\end{tabular}
\end{table}

\subsection{PDEZoo}
\label{subsec:PDEZoo_problem_append}
Instances use $\mathcal P=(\Omega,\mathcal E,\vartheta,f,\mathcal G,u)$ from Sec.~\ref{subsec:PDEZoo_problem}, with $\vartheta=\lambda$ for Yukawa, $k^2$ for Helmholtz, and empty otherwise, and boundary data $\mathcal G=g$ for second-order equations or $(g_1,g_2)$ for biharmonic. PDEZoo includes five elliptic PDE families:

\textbf{Laplace.}
The solution satisfies $\Delta u = 0 \:\: \text{in} \:\: \Omega,$
with Dirichlet boundary condition $u = g \:\: \text{on} \:\: \partial \Omega.$
In this setting, $u$ is constructed from analytically harmonic atoms.

\textbf{Poisson.}
The solution satisfies $\Delta u = f \:\: \text{in} \:\: \Omega,$
with Dirichlet boundary values inherited from the manufactured solution. Here, $u$ is sampled from a rich family of smooth non-harmonic expressions, and the source term is obtained as $f = \Delta u$. 

\textbf{Yukawa.}
The solution satisfies $\Delta u - \lambda u = f \:\: \text{in} \:\: \Omega,$
with screening parameter $\lambda > 0$. The source is defined from the manufactured solution by $f = \Delta u - \lambda u$. In our dataset, $\lambda$ is sampled uniformly from $[0.5, 50.0]$.

\textbf{Biharmonic.}
The solution satisfies $\Delta^2u=f$ in $\Omega$, with Navier data $u=g_1$ and $\Delta u=g_2$ on $\partial\Omega$, where $(g_1,g_2)=(u,\Delta u)|_{\partial\Omega}$.
Here, $u$ is sampled from a family of smooth manufactured expressions, and the source term is obtained as $f = \Delta^2 u$, where $\Delta^2 = \Delta(\Delta \cdot)$ denotes the biharmonic operator. Setting $v=\Delta u$ gives $\Delta v=f$, $v|_{\partial\Omega}=g_2$, followed by $\Delta u=v$, $u|_{\partial\Omega}=g_1$.

\textbf{Helmholtz.}
The solution satisfies $\Delta u + k^2 u = f \:\: \text{in} \:\: \Omega,$
with wavenumber $k > 0$ and Dirichlet boundary values inherited from the manufactured solution. The source is defined as $f = \Delta u + k^2 u$. A body-fitted quadratic-FEM check of all 1,000 Helmholtz domains, refined near the minima, gives $\max k^2=1.999$, $\min\lambda_1\simeq3.773$, and $\min(\lambda_1/k^2)\simeq2.140$, where $\lambda_1$ is the first Dirichlet eigenvalue of $-\Delta$. These instances lie below the first resonance. The generator samples $k^2\sim\mathcal U[0.1,2.0]$.
\begin{table}[t]
\centering
\footnotesize
\caption{WoS MSE slopes over budgets 1--32 on 501 baseline-generator diagnostic cases, distinct from the headline test split. Variance-dominated $1/K$ scaling has slope $-1$.}
\label{tab:convergence}
\begin{tabular}{@{}lcccc@{}}
\toprule
PDE Type & Median Slope & Min & Max & \% in $[-1.05, -0.95]$ \\
\midrule
Laplace & $-1.000$ & $-1.015$ & $-0.981$ & 100.0\% \\
Poisson & $-1.000$ & $-1.045$ & $-0.918$ & 99.4\% \\
Yukawa & $-1.000$ & $-1.043$ & $-0.944$ & 99.4\% \\
\midrule
\textbf{All} & $\mathbf{-1.000}$ & $\mathbf{-1.045}$ & $\mathbf{-0.918}$ & $\mathbf{99.6\%}$ \\
\bottomrule
\end{tabular}
\end{table}

\section{Monte Carlo Walk-On-Spheres Solver}

\subsection{WoS Solver Parameters}
\label{app:solver}

\paragraph{Ball estimator and held-out families.}
For independent uniform $Z\in\partial B_R(x)$ and $Y\in B_R(x)$, our source convention gives
\[
u(x)=\psi(R)\mathbb E[u(Z)]-\pi R^2\mathbb E[G(|Y-x|,R)f(Y)].
\]
Poisson uses $\psi=1$, $G=(2\pi)^{-1}\log(R/r)$, and Laplace sets $f=0$. With $\mu=\sqrt\lambda$, Yukawa uses $\psi=I_0(\mu R)^{-1}$ and
\[
G=(2\pi)^{-1}[K_0(\mu r)-K_0(\mu R)I_0(\mu r)/I_0(\mu R)].
\]
Helmholtz uses
\[
\psi=J_0(kR)^{-1},\quad
G=\frac{Y_0(kR)J_0(kr)-J_0(kR)Y_0(kr)}{4J_0(kR)}.
\]
Here $I_0,K_0,J_0,Y_0$ are Bessel functions. The Helmholtz Green function satisfies $(\Delta+k^2)G=-\delta_x$, $G(R)=0$, and $\partial_rG(R)=-1/(2\pi R J_0(kR))$, yielding the boundary factor above. The radius $R\le\min(d_{\rm safe},2/k)$ avoids a singular single-step weight. 

Initialize $X=x$ and $(A,w)=(0,1)$. At each step sample $Y,Z$, accumulate $A\leftarrow A-w\pi R^2Gf(Y)$, update $w\leftarrow w\psi$, and set $X\leftarrow Z$. Return $A+w g(\Pi_{\partial\Omega}X)$ at shell/cap termination. Uniform disk sampling uses radius $R\sqrt U$, $U\sim\mathcal U[0,1]$. For biharmonic, replace the outer source $v(Y)$ by one independent inner Poisson estimate with source $f$ and boundary $g_2$. The tower property justifies this only when inner estimates are unbiased and integrable.

The Walk-on-Spheres solver uses the following parameters:

\begin{table}[h]
\centering
\footnotesize
\caption{WoS solver parameters.}
\label{tab:solver_params}
\begin{tabular}{@{}llp{6cm}@{}}
\toprule
Parameter & Value & Description \\
\midrule
Resolution & $256 \times 256$ & Uniform grid over $[-1,1]^2$ \\
$\varepsilon$ (shell) & $10^{-4}$ & Boundary termination threshold \\
Outer walk cap & 128 or 256 & Generation-run dependent; per-case metadata are authoritative \\
Biharmonic inner cap & 16 & Used in 985 of 994 audited cases \\
Held-out source grid & $1024\times1024$ & Bilinear lookup: Helmholtz $f$; biharmonic $f,\Delta u$ \\
walk\_batch & 256 & Walks processed per GPU batch \\
\bottomrule
\end{tabular}
\end{table}

\textbf{Numerical Bias} 
The solver uses shell threshold $\varepsilon=10^{-4}$. Shell termination, finite-cap boundary projection, and source interpolation can introduce combined bias $b_\eta$, for which we do not assert a universal $\varepsilon$-only rate. In a biharmonic audit of 1000 cases, 99\% of examples used a 16-step inner cap, 27--33\% of audited two-dimensional walks remained unfinished at 16 steps, and the raw high-budget error floor was near 0.8\%, which more samples or outer-cap sweeps with a fixed inner cap do not remove.

\section{MC$^2$ Architecture}

Our MC$^2$ corrector is instantiated as a Swin-UNet (SUNet) architecture \cite{Fan_2022}. The model processes $256 \times 256$ inputs with patch size 4, embedding dimension 96, and four encoder stages with depths $[8, 8, 8, 8]$ and $[8, 8, 8, 8]$ attention heads. The decoder mirrors the encoder with skip connections. We use a window size of 8 for shifted-window self-attention, MLP ratio 4, and stochastic depth with drop path rate 0.1. The final layer uses dual upsampling to reconstruct the full-resolution output. The two-channel model has 99,681,129 parameters and a global residual connection. We train for 40 epochs using AdamW, cosine learning-rate decay from $2\times10^{-4}$ toward $10^{-6}$, FP32, and batch size 24/GPU (96 globally).  Training is distributed across 4 GPUs using PyTorch's DDP. Training minimizes masked $L^1$ error in the normalized frame. The conditional-mean/MSE analysis is therefore a theoretical benchmark, not the exact training objective.  We select the checkpoint with the best validation PSNR. 

\section{Baseline Models}

\subsection{Denoisers}

We use Gaussian $\sigma=1$, heavy Gaussian $\sigma=3$, TV weight $0.1$, and NLM patch size 5, search distance 6, $h=0.1$. Median and bilateral filters are not in the reported panel. 

\subsection{FNO}

The architecture uses 4 Fourier layers with 16 modes and width 64. The retained checkpoint recipe uses 30 epochs, masked relative-$L^2$ loss and cosine scheduling. Reconstructed commands specify Adam, base LR $10^{-3}$ and batch 16/GPU on four GPUs, with the code scaling LR by world size. Some original commands were reconstructed rather than recovered from logs.

The matched FNO corrector uses width 156, 99,799,229 parameters and $(\hat u_8,f)$, and the U-Net uses width 114 and 99,559,507 parameters. FNO uses its 30-epoch recipe, and U-Net follows the single-budget B8 schedule, while headline SUNet is multi-budget.

\textcolor{black}{The Laplace diffusion teacher uses run \texttt{00018} at 512k training images. Poisson and Yukawa teachers are retrained using the original recipe, with checkpoints at 153k and 122k images, respectively. Checkpoints are selected using training loss and sample statistics without test data. Both modes use 100-step Heun sampling with $\sigma_{\max}=80$, $\sigma_{\min}=0.002$, and $\rho=7$. Conditioning inpaints $(f,\lambda,m)$, with boundary values $g$ additionally inpainted on a one-pixel boundary shell, and uses the $B=8$ WoS field as a noisy observation $y=u+\eta$. The per-pixel noise level $s(x)$ is estimated from $y$ alone, and RePaint-style replacement is applied while $\sigma_t>s(x)$. DiffusionPDE-mode applies physics guidance during the final $20\%$ of sampling, with coefficients calibrated on training samples: $2.48\times10^{-6}$ for Laplace, $1.56\times10^{-4}$ for Poisson, and $1.37\times10^{-4}$ for Yukawa. CoCoGen-mode adds ten physics-only steps using the same calibrated coefficient. These modes share teachers and are adaptations rather than independent reproductions of the original methods.}

\section{Extended Results}

\subsection{Full Per-Family Results}
\label{app:full_per_family} \begin{table}[t] \centering \small \caption{\textcolor{black}{Median raw relative $L^2$ error on the Laplace, Poisson, and Yukawa cases. Both diffusion modes use $B=8$ WoS observations and boundary values.}} \label{tab:diffusion_relative_l2} \begin{tabular}{lrrrr} \toprule \textcolor{black}{Method} & \textcolor{black}{Laplace} & \textcolor{black}{Poisson} & \textcolor{black}{Yukawa} & \textcolor{black}{Pooled} \\ \midrule \textcolor{black}{MC$^2$} & \textcolor{black}{0.0097} & \textcolor{black}{0.0272} & \textcolor{black}{0.0144} & \textcolor{black}{0.0132} \\  \textcolor{black}{DiffusionPDE-mode} & \textcolor{black}{0.133} & \textcolor{black}{0.175} & \textcolor{black}{0.099} & \textcolor{black}{0.133} \\  \textcolor{black}{CoCoGen-mode} & \textcolor{black}{0.164} & \textcolor{black}{0.175} & \textcolor{black}{0.099} & \textcolor{black}{0.143} \\  \textcolor{black}{WoS ($B=8$)} & \textcolor{black}{0.536} & \textcolor{black}{2.55} & \textcolor{black}{0.945} & \textcolor{black}{0.785} \\  \bottomrule \end{tabular} \end{table} \par 
For completeness, Table~\ref{tab:method_comparison_full} reports per-family results across all four metrics (PSNR, SNR, MSE, LPIPS) for the original baseline panel and headline corrector, including the full Walk-on-Spheres compute curve from $B\!=\!8$ to $B\!=\!131{,}072$.  The aggregate trends mirror the main-paper findings: MC$^2$ from a single $B\!=\!8$ WoS input matches WoS quality at $B\!\sim\!16{,}384$--$32{,}768$ on in-distribution families and at $B\!\sim\!8{,}192$--$16{,}384$ on out-of-distribution families, corresponding to a $1{,}024$--$4{,}096\times$ reduction in Monte Carlo compute.  The per-family breakdown also reveals where each baseline method's weaknesses concentrate. PINO's physics regularization actively reduces performance on every family, producing solutions worse than the raw $B\!=\!8$ input. \textcolor{black}{With $B=8$ WoS observations and boundary values, CoCoGen-mode reaches $30.07$ dB pooled mean PSNR and DiffusionPDE-mode reaches $30.42$ dB on the three in-distribution families. MC$^2$ reaches $49.55$ dB on the same family group. On the evaluated cases, median raw relative $L^2$ error is $0.143$ for CoCoGen-mode, $0.133$ for DiffusionPDE-mode, and $0.0132$ for MC$^2$.} The matched U-Net/FNO correctors achieve $37.32/31.25$ dB pooled mean PSNR versus MC$^2$ at $44.90$ dB.  Among traditional denoisers, Non-Local Means and Total Variation provide the largest gains over raw WoS, but remain below the headline corrector.  LPIPS values reinforce the PSNR ranking: MC$^2$ is the only method that achieves perceptually-near-zero error ($<0.063$) on every family.

\begin{table*}[p]
\centering
\caption{Full per-family results across all four metrics on the PDEZoo test set. \textcolor{black}{Diffusion rows use $B=8$ WoS observations and boundary values, and their overall values pool only the Laplace, Poisson, and Yukawa cases.} PSNR and SNR are reported in dB; MSE and LPIPS are unitless. Arrows indicate metric direction. \textbf{Bold} marks the best result per column among learned and post-processing methods. MC$^2$ uses a single $B\!=\!8$ WoS input.}
\label{tab:method_comparison_full}

\setlength{\tabcolsep}{0.6pt}
\renewcommand{\arraystretch}{1.9}
\small
\begin{sc}
\begin{adjustbox}{angle=90,width=\textwidth,max totalheight=0.87\textheight}
\begin{tabular}{@{}l cccc cccc cccc cccc cccc cccc@{}}
\toprule
& \multicolumn{4}{c}{Laplace}
& \multicolumn{4}{c}{Poisson}
& \multicolumn{4}{c}{Yukawa}
& \multicolumn{4}{c}{Biharmonic}
& \multicolumn{4}{c}{Helmholtz}
& \multicolumn{4}{c}{\textbf{Overall}} \\
\cmidrule(lr){2-5}
\cmidrule(lr){6-9}
\cmidrule(lr){10-13}
\cmidrule(lr){14-17}
\cmidrule(lr){18-21}
\cmidrule(lr){22-25}
Method
& PSNR$\uparrow$ & SNR$\uparrow$ & MSE$\downarrow$ & LPIPS$\downarrow$
& PSNR$\uparrow$ & SNR$\uparrow$ & MSE$\downarrow$ & LPIPS$\downarrow$
& PSNR$\uparrow$ & SNR$\uparrow$ & MSE$\downarrow$ & LPIPS$\downarrow$
& PSNR$\uparrow$ & SNR$\uparrow$ & MSE$\downarrow$ & LPIPS$\downarrow$
& PSNR$\uparrow$ & SNR$\uparrow$ & MSE$\downarrow$ & LPIPS$\downarrow$
& PSNR$\uparrow$ & SNR$\uparrow$ & MSE$\downarrow$ & LPIPS$\downarrow$ \\
\midrule

\multicolumn{25}{l}{\textit{Walk-on-Spheres (Monte Carlo)}} \\

WoS ($B\!=\!8$)
& 25.22 & 19.17 & 4.02e--03 & 0.5188
& 11.20 & 5.90 & 0.1214 & 0.8220
& 13.89 & 8.48 & 0.0793 & 0.7298
& 15.01 & 9.65 & 0.0716 & 0.7315
& 10.93 & 5.60 & 0.1247 & 0.8360
& 15.26 & 9.77 & 0.0802 & 0.7275 \\

WoS ($B\!=\!16$)
& 28.22 & 22.17 & 2.07e--03 & 0.4377
& 12.41 & 7.11 & 0.1086 & 0.7957
& 15.56 & 10.15 & 0.0652 & 0.6886
& 16.61 & 11.25 & 0.0665 & 0.6958
& 12.06 & 6.73 & 0.1126 & 0.8131
& 16.99 & 11.50 & 0.0710 & 0.6861 \\

WoS ($B\!=\!32$)
& 31.21 & 25.17 & 1.06e--03 & 0.3561
& 13.89 & 8.59 & 0.0936 & 0.7627
& 17.50 & 12.09 & 0.0511 & 0.6416
& 18.42 & 13.05 & 0.0603 & 0.6553
& 13.48 & 8.14 & 0.0982 & 0.7831
& 18.91 & 13.42 & 0.0609 & 0.6396 \\

WoS ($B\!=\!64$)
& 34.22 & 28.17 & 5.35e--04 & 0.2782
& 15.63 & 10.33 & 0.0777 & 0.7229
& 19.67 & 14.26 & 0.0383 & 0.5901
& 20.40 & 15.04 & 0.0533 & 0.6107
& 15.15 & 9.82 & 0.0828 & 0.7451
& 21.03 & 15.54 & 0.0506 & 0.5893 \\

WoS ($B\!=\!128$)
& 37.23 & 31.18 & 2.70e--04 & 0.2090
& 17.60 & 12.30 & 0.0622 & 0.6772
& 22.05 & 16.64 & 0.0273 & 0.5336
& 22.54 & 17.18 & 0.0457 & 0.5640
& 17.06 & 11.73 & 0.0674 & 0.7015
& 23.31 & 17.82 & 0.0407 & 0.5370 \\

WoS ($B\!=\!256$)
& 40.23 & 34.19 & 1.35e--04 & 0.1520
& 19.78 & 14.47 & 0.0479 & 0.6273
& 24.60 & 19.19 & 0.0185 & 0.4725
& 24.82 & 19.46 & 0.0381 & 0.5161
& 19.18 & 13.85 & 0.0530 & 0.6529
& 25.74 & 20.24 & 0.0316 & 0.4841 \\

WoS ($B\!=\!512$)
& 43.25 & 37.20 & 6.78e--05 & 0.1065
& 22.13 & 16.83 & 0.0353 & 0.5733
& 27.28 & 21.87 & 0.0119 & 0.4094
& 27.21 & 21.85 & 0.0306 & 0.4676
& 21.49 & 16.16 & 0.0402 & 0.6005
& 28.29 & 22.79 & 0.0237 & 0.4315 \\

WoS ($B\!=\!1024$)
& 46.26 & 40.21 & 3.39e--05 & 0.0700
& 24.65 & 19.35 & 0.0248 & 0.5161
& 30.07 & 24.66 & 7.28e--03 & 0.3489
& 29.68 & 24.31 & 0.0238 & 0.4199
& 23.95 & 18.62 & 0.0294 & 0.5446
& 30.94 & 25.44 & 0.0171 & 0.3800 \\

WoS ($B\!=\!2048$)
& 49.27 & 43.22 & 1.70e--05 & 0.0419
& 27.31 & 22.01 & 0.0165 & 0.4560
& 32.95 & 27.54 & 4.25e--03 & 0.2924
& 32.20 & 26.84 & 0.0178 & 0.3717
& 26.56 & 21.22 & 0.0206 & 0.4866
& 33.67 & 28.17 & 0.0119 & 0.3299 \\

WoS ($B\!=\!4096$)
& 52.27 & 46.23 & 8.49e--06 & 0.0228
& 30.08 & 24.78 & 0.0104 & 0.3954
& 35.87 & 30.46 & 2.38e--03 & 0.2399
& 34.77 & 29.40 & 0.0127 & 0.3241
& 29.26 & 23.93 & 0.0138 & 0.4270
& 36.46 & 30.97 & 7.91e--03 & 0.2820 \\

WoS ($B\!=\!8192$)
& 55.29 & 49.24 & 4.25e--06 & 0.0111
& 32.94 & 27.64 & 6.26e--03 & 0.3367
& 38.83 & 33.42 & 1.28e--03 & 0.1902
& 37.30 & 31.93 & 8.75e--03 & 0.2799
& 32.05 & 26.72 & 8.91e--03 & 0.3681
& 39.29 & 33.80 & 5.07e--03 & 0.2375 \\

WoS ($B\!=\!16384$)
& 58.30 & 52.25 & 2.12e--06 & 0.0049
& 35.86 & 30.55 & 3.57e--03 & 0.2824
& 41.81 & 36.41 & 6.73e--04 & 0.1431
& 39.77 & 34.41 & 5.84e--03 & 0.2384
& 34.89 & 29.56 & 5.49e--03 & 0.3115
& 42.14 & 36.64 & 3.13e--03 & 0.1964 \\

WoS ($B\!=\!32768$)
& 61.31 & 55.26 & 1.06e--06 & 0.0019
& 38.81 & 33.51 & 1.96e--03 & 0.2329
& 44.80 & 39.39 & 3.45e--04 & 0.1006
& 42.15 & 36.79 & 3.80e--03 & 0.2004
& 37.77 & 32.44 & 3.25e--03 & 0.2583
& 44.98 & 39.49 & 1.88e--03 & 0.1592 \\

WoS ($B\!=\!65536$)
& 64.31 & 58.27 & 5.31e--07 & 0.0007
& 41.78 & 36.47 & 1.04e--03 & 0.1874
& 47.79 & 42.38 & 1.75e--04 & 0.0654
& 44.40 & 39.03 & 2.44e--03 & 0.1656
& 40.67 & 35.34 & 1.87e--03 & 0.2102
& 47.80 & 42.31 & 1.11e--03 & 0.1263 \\

WoS ($B\!=\!131072$)
& 67.32 & 61.27 & 2.66e--07 & 0.0003
& 44.77 & 39.47 & 5.40e--04 & 0.1442
& 50.80 & 45.38 & 8.83e--05 & 0.0390
& 46.49 & 41.13 & 1.57e--03 & 0.1347
& 43.56 & 38.23 & 1.06e--03 & 0.1669
& 50.60 & 45.10 & 6.56e--04 & 0.0974 \\

\midrule
\multicolumn{25}{l}{\textit{Traditional Denoisers (applied to $B\!=\!8$ WoS)}} \\

Gaussian
& 29.67 & 23.62 & $1.30\times10^{-3}$ & 0.2755
& 14.64 & 9.34 & $7.63\times10^{-2}$ & 0.4625
& 17.57 & 12.16 & $3.91\times10^{-2}$ & 0.4045
& 18.11 & 12.74 & $5.45\times10^{-2}$ & 0.4008
& 14.21 & 8.88 & $8.14\times10^{-2}$ & 0.4730
& 18.85 & 13.36 & $5.06\times10^{-2}$ & 0.4032 \\

Gaussian (heavy)
& 25.98 & 19.93 & $3.30\times10^{-3}$ & 0.1641
& 14.64 & 9.34 & $5.63\times10^{-2}$ & 0.3308
& 16.55 & 11.14 & $3.46\times10^{-2}$ & 0.2459
& 16.73 & 11.36 & $4.76\times10^{-2}$ & 0.3269
& 14.32 & 8.99 & $6.12\times10^{-2}$ & 0.3473
& 17.66 & 12.16 & $4.06\times10^{-2}$ & 0.2832 \\

Total Variation
& 37.81 & 31.76 & $2.07\times10^{-4}$ & 0.1182
& 19.34 & 14.04 & $4.29\times10^{-2}$ & 0.3506
& 23.34 & 17.93 & $1.64\times10^{-2}$ & 0.2153
& 23.81 & 18.44 & $3.04\times10^{-2}$ & 0.2747
& 18.80 & 13.46 & $4.51\times10^{-2}$ & 0.3462
& 24.63 & 19.14 & $2.71\times10^{-2}$ & 0.2613 \\

Non-Local Means
& 39.16 & 33.11 & $1.57\times10^{-4}$ & 0.0959
& 21.56 & 16.26 & $2.95\times10^{-2}$ & 0.4277
& 25.54 & 20.13 & $1.03\times10^{-2}$ & 0.3152
& 25.91 & 20.55 & $2.32\times10^{-2}$ & 0.3258
& 20.88 & 15.55 & $3.29\times10^{-2}$ & 0.4415
& 26.63 & 21.13 & $1.93\times10^{-2}$ & 0.3212 \\

\midrule
\multicolumn{25}{l}{\textit{Neural Network Methods}} \\

FNO operator ($f$)
& 16.52 & 10.47 & $5.95\times10^{-2}$ & 0.1959
& 5.47 & 0.17 & $3.07\times10^{-1}$ & 0.4546
& 5.68 & 0.27 & $2.94\times10^{-1}$ & 0.4410
& 5.43 & 0.07 & $3.25\times10^{-1}$ & 0.3519
& 5.52 & 0.18 & $3.07\times10^{-1}$ & 0.4541
& 7.74 & 2.25 & $2.58\times10^{-1}$ & 0.3790 \\

PINO operator ($f$)
& 18.46 & 12.41 & $4.69\times10^{-2}$ & 0.1810
& 5.71 & 0.40 & $2.93\times10^{-1}$ & 0.4492
& 5.65 & 0.24 & $2.95\times10^{-1}$ & 0.4379
& 5.54 & 0.18 & $3.15\times10^{-1}$ & 0.3619
& 5.70 & 0.36 & $2.97\times10^{-1}$ & 0.4489
& 8.24 & 2.75 & $2.49\times10^{-1}$ & 0.3753 \\

CoCoGen-mode
& \textcolor{black}{33.31} & \textcolor{black}{27.261} & \textcolor{black}{$1.25\times10^{-2}$} & \textcolor{black}{0.1386}
& \textcolor{black}{27.11} & \textcolor{black}{21.805} & \textcolor{black}{$9.34\times10^{-3}$} & \textcolor{black}{0.0511}
& \textcolor{black}{29.76} & \textcolor{black}{24.345} & \textcolor{black}{$4.45\times10^{-3}$} & \textcolor{black}{0.0494}
& -- & -- & -- & --
& -- & -- & -- & --
& \textcolor{black}{30.07} & \textcolor{black}{24.476} & \textcolor{black}{$8.80\times10^{-3}$} & \textcolor{black}{0.0801} \\

DiffusionPDE-mode
& \textcolor{black}{34.40} & \textcolor{black}{28.355} & \textcolor{black}{$8.49\times10^{-3}$} & \textcolor{black}{0.1809}
& \textcolor{black}{27.12} & \textcolor{black}{21.822} & \textcolor{black}{$9.34\times10^{-3}$} & \textcolor{black}{0.0528}
& \textcolor{black}{29.70} & \textcolor{black}{24.293} & \textcolor{black}{$4.45\times10^{-3}$} & \textcolor{black}{0.0546}
& -- & -- & -- & --
& -- & -- & -- & --
& \textcolor{black}{30.42} & \textcolor{black}{24.835} & \textcolor{black}{$7.46\times10^{-3}$} & \textcolor{black}{0.0967} \\

\midrule
\textbf{MC$^2$ (Ours)}
& \textbf{59.94} & \textbf{53.89} & $\mathbf{1.67\times10^{-6}}$ & \textbf{0.0001}
& \textbf{42.17} & \textbf{36.86} & $\mathbf{5.41\times10^{-4}}$ & \textbf{0.0056}
& \textbf{46.40} & \textbf{40.99} & $\mathbf{1.32\times10^{-4}}$ & \textbf{0.0015}
& \textbf{35.96} & \textbf{30.60} & $\mathbf{7.89\times10^{-3}}$ & \textbf{0.0633}
& \textbf{40.05} & \textbf{34.72} & $\mathbf{1.08\times10^{-3}}$ & \textbf{0.0110}
& \textbf{44.90} & \textbf{39.41} & $\mathbf{1.94\times10^{-3}}$ & \textbf{0.0164} \\

\bottomrule
\end{tabular}
\end{adjustbox}
\end{sc}
\end{table*}   
\begin{table*}[t]
\centering
\caption{Effect of training data scale on MC$^2$ performance. All models use $B\!=\!8$ WoS inputs. Fractions denote the proportion of the full training set used.} 
\label{tab:mc2_data_scaling}
\setlength{\tabcolsep}{2.2pt}
\renewcommand{\arraystretch}{1.05}
\footnotesize
\resizebox{\textwidth}{!}{%
\begin{tabular}{@{}l cccc cccc cccc cccc cccc cccc@{}}
\toprule
& \multicolumn{4}{c}{Biharmonic} & \multicolumn{4}{c}{Helmholtz} & \multicolumn{4}{c}{Laplace} & \multicolumn{4}{c}{Poisson} & \multicolumn{4}{c}{Yukawa} & \multicolumn{4}{c}{\textbf{Overall}} \\
\cmidrule(lr){2-5} \cmidrule(lr){6-9} \cmidrule(lr){10-13} \cmidrule(lr){14-17} \cmidrule(lr){18-21} \cmidrule(lr){22-25}
Data fraction
& PSNR$\uparrow$ & SNR$\uparrow$ & MSE$\downarrow$ & LPIPS$\downarrow$
& PSNR$\uparrow$ & SNR$\uparrow$ & MSE$\downarrow$ & LPIPS$\downarrow$
& PSNR$\uparrow$ & SNR$\uparrow$ & MSE$\downarrow$ & LPIPS$\downarrow$
& PSNR$\uparrow$ & SNR$\uparrow$ & MSE$\downarrow$ & LPIPS$\downarrow$
& PSNR$\uparrow$ & SNR$\uparrow$ & MSE$\downarrow$ & LPIPS$\downarrow$
& PSNR$\uparrow$ & SNR$\uparrow$ & MSE$\downarrow$ & LPIPS$\downarrow$ \\
\midrule
0.04$\times$ & 31.93 & 26.57 & $1.38\times10^{-2}$ & 0.2060 & 30.47 & 25.13 & $5.73\times10^{-3}$ & 0.2427 & 50.63 & 44.59 & $1.16\times10^{-5}$ & 0.0044 & 31.65 & 26.35 & $3.51\times10^{-3}$ & 0.2316 & 35.98 & 30.57 & $1.11\times10^{-3}$ & 0.1030 & 36.14 & 30.65 & $4.85\times10^{-3}$ & 0.1579 \\

0.08$\times$ & 33.79 & 28.43 & $1.17\times10^{-2}$ & 0.1615 & 34.36 & 29.03 & $3.10\times10^{-3}$ & 0.1374 & 53.74 & 47.69 & $5.89\times10^{-6}$ & 0.0009 & 35.73 & 30.43 & $1.58\times10^{-3}$ & 0.1233 & 40.08 & 34.67 & $4.50\times10^{-4}$ & 0.0361 & 39.55 & 34.05 & $3.38\times10^{-3}$ & 0.0922 \\

0.16$\times$ & 34.26 & 28.89 & $1.28\times10^{-2}$ & 0.1321 & 35.06 & 29.73 & $3.31\times10^{-3}$ & 0.0819 & 54.70 & 48.65 & $4.83\times10^{-6}$ & 0.0004 & 36.54 & 31.24 & $1.54\times10^{-3}$ & 0.0686 & 40.84 & 35.43 & $4.60\times10^{-4}$ & 0.0163 & 40.28 & 34.79 & $3.65\times10^{-3}$ & 0.0601 \\

0.32$\times$ & 35.70 & 30.34 & $9.29\times10^{-3}$ & 0.1406 & 35.63 & 30.29 & $3.16\times10^{-3}$ & 0.1127 & 55.38 & 49.33 & $4.05\times10^{-6}$ & 0.0004 & 37.13 & 31.83 & $1.52\times10^{-3}$ & 0.0857 & 41.13 & 35.72 & $4.74\times10^{-4}$ & 0.0222 & 41.00 & 35.51 & $2.90\times10^{-3}$ & 0.0726 \\

0.64$\times$ & \textbf{36.06} & \textbf{30.70} & $9.23\times10^{-3}$ & 0.0847 & 38.74 & 33.40 & $1.43\times10^{-3}$ & 0.0188 & 58.67 & 52.62 & $2.11\times10^{-6}$ & \textbf{0.0001} & 40.76 & 35.46 & $5.62\times10^{-4}$ & 0.0113 & 45.13 & 39.71 & $1.33\times10^{-4}$ & 0.0025 & 43.87 & 38.38 & $2.28\times10^{-3}$ & 0.0236 \\

\midrule
1.0$\times$ (single-budget) & 35.55 & 30.19 & $9.56\times10^{-3}$ & 0.0691 & 39.77 & 34.44 & $1.17\times10^{-3}$ & 0.0111 & \textbf{60.07} & \textbf{54.03} & $\mathbf{1.55\times10^{-6}}$ & \textbf{0.0001} & \textbf{42.59} & \textbf{37.29} & $\mathbf{3.78\times10^{-4}}$ & \textbf{0.0048} & \textbf{46.86} & \textbf{41.45} & $\mathbf{9.40\times10^{-5}}$ & \textbf{0.0014} & \textbf{44.96} & \textbf{39.47} & $2.25\times10^{-3}$ & 0.0174 \\

Headline (multi-budget) & 35.96 & 30.60 & $\mathbf{7.89\times10^{-3}}$ & \textbf{0.0633} & \textbf{40.05} & \textbf{34.72} & $\mathbf{1.08\times10^{-3}}$ & \textbf{0.0110} & 59.94 & 53.89 & $1.67\times10^{-6}$ & \textbf{0.0001} & 42.17 & 36.86 & $5.41\times10^{-4}$ & 0.0056 & 46.40 & 40.99 & $1.32\times10^{-4}$ & 0.0015 & 44.90 & 39.41 & $\mathbf{1.94\times10^{-3}}$ & \textbf{0.0164} \\

\bottomrule
\end{tabular}%
}
\end{table*}

\subsection{Controlled input ablation}\label{app:input_ablation}
Four identical 99,682,857-parameter SUNets use B=8 data and 40 epochs. On the test set, median PSNR for $(\hat u,f,m)/(\hat u,m)/(f,m)/(f,g,m)$ is $44.29/40.56/9.29/8.56$ dB. Mean PSNR with/without $f$ is Laplace $57.79/57.77$, Poisson $39.79/33.25$, Yukawa $44.23/37.66$, biharmonic $36.27/38.96$, and Helmholtz $37.98/32.45$. Fixed B=8-trained observation-bearing arms have median raw relative $L^2$ $0.056/0.065$ at B2, $0.021/0.029$ at B8, and approximately $0.013/0.013$ at B4,096. 

\paragraph{Source conditioning across input budgets.}
We evaluate the fixed $B=8$-trained correctors with inputs $(\hat u_B,f,m)$ and $(\hat u_B,m)$ at all 17 stored budgets, without retraining. Tables~\ref{tab:input_budget_psnr} and~\ref{tab:input_budget_l2} report medians on paired finite cases. At $B=8$, source conditioning raises pooled median PSNR from $40.56$ to $44.29$ dB and reduces median raw relative $L^2$ error from $0.0289$ to $0.0207$. Its effect varies across PDE families. At this budget, Poisson, Yukawa, and Helmholtz gain $6.45$, $6.55$, and $5.69$ dB, respectively, while Laplace changes little and biharmonic loses $3.18$ dB. The pooled relative-error gap largely disappears at high budgets, although family-specific differences remain. These results support source conditioning for several families at low compute, rather than a uniform benefit across operators and budgets.

\paragraph{Reporting protocol.}
Each entry lists the source-conditioned / no-source median. PSNR uses the benchmark's normalized, clipped metric, whereas relative $L^2$ uses the original fields without clipping. Pairing is performed separately for each metric, family, and budget. The two metrics have identical paired counts in this evaluation. Laplace and Helmholtz each have $n=1000$ at every budget. Poisson has $n=996$ through $B=32768$ and $n=995$ thereafter. Biharmonic has $n=995$ through $B=2048$ and $n=994$ thereafter. Yukawa has $n=960$ through $B=2048$, $n=959$ from $B=4096$ through $32768$, $n=958$ at $65536$, and $n=957$ at $131072$. Pooled counts are therefore $4951$, $4949$, $4947$, and $4946$ over these respective budget ranges. Pooled medians are computed over cases, not averaged over family medians. The accompanying \texttt{paired\_budget\_summary.csv} provides both medians and per-metric counts for every entry.

\begin{table*}[t]
\centering
\small
\setlength{\tabcolsep}{5pt}
\caption{Median MC$^2$ PSNR (dB, higher is better) for fixed $(B=8)$-trained input-ablation models evaluated at budget $B$. Entries give source-conditioned / no-source results. Counts and pairing are specified in the reporting protocol.}
\label{tab:input_budget_psnr}
\begin{tabular}{rcccccc}
\toprule
$B$ & Laplace & Poisson & Yukawa & Biharmonic & Helmholtz & Pooled \\
\midrule
2 & $48.00 / 47.86$ & $32.30 / 27.75$ & $37.61 / 33.42$ & $33.28 / 36.51$ & $30.94 / 27.23$ & $37.82 / 35.35$ \\
4 & $54.62 / 54.72$ & $36.16 / 30.45$ & $41.35 / 36.08$ & $35.80 / 38.68$ & $34.53 / 29.59$ & $41.47 / 38.09$ \\
8 & $58.17 / 58.10$ & $39.35 / 32.89$ & $44.73 / 38.18$ & $38.15 / 41.33$ & $37.62 / 31.94$ & $44.29 / 40.56$ \\
16 & $58.53 / 58.45$ & $40.91 / 34.79$ & $46.59 / 39.91$ & $40.42 / 43.30$ & $39.90 / 33.92$ & $46.20 / 42.50$ \\
32 & $58.22 / 58.40$ & $42.13 / 36.31$ & $47.02 / 41.46$ & $42.21 / 44.80$ & $41.46 / 35.59$ & $47.63 / 44.36$ \\
64 & $58.19 / 58.53$ & $43.01 / 37.79$ & $47.62 / 42.21$ & $43.46 / 46.11$ & $41.95 / 36.93$ & $48.67 / 45.76$ \\
128 & $58.30 / 58.87$ & $44.15 / 39.08$ & $47.66 / 43.03$ & $44.83 / 47.28$ & $42.81 / 38.24$ & $49.49 / 46.92$ \\
256 & $58.44 / 59.14$ & $44.34 / 40.06$ & $47.60 / 43.66$ & $45.57 / 48.33$ & $43.46 / 39.21$ & $50.04 / 47.87$ \\
512 & $58.47 / 59.37$ & $44.86 / 40.97$ & $47.49 / 43.81$ & $46.32 / 49.42$ & $44.14 / 40.16$ & $50.53 / 48.85$ \\
1024 & $58.48 / 59.40$ & $45.30 / 41.77$ & $47.44 / 44.05$ & $46.98 / 50.13$ & $44.88 / 41.10$ & $50.88 / 49.66$ \\
2048 & $58.47 / 59.44$ & $45.62 / 42.30$ & $47.28 / 44.20$ & $47.64 / 50.80$ & $44.83 / 41.55$ & $51.06 / 50.23$ \\
4096 & $58.54 / 59.45$ & $45.71 / 42.74$ & $47.19 / 44.34$ & $48.09 / 51.51$ & $45.17 / 42.16$ & $51.30 / 50.55$ \\
8192 & $58.52 / 59.42$ & $45.67 / 43.06$ & $47.05 / 44.29$ & $48.48 / 52.20$ & $45.35 / 42.57$ & $51.42 / 50.91$ \\
16384 & $58.47 / 59.41$ & $45.54 / 43.24$ & $47.01 / 44.34$ & $48.85 / 52.62$ & $45.66 / 42.87$ & $51.50 / 51.18$ \\
32768 & $58.46 / 59.42$ & $45.62 / 43.42$ & $46.99 / 44.46$ & $49.24 / 53.25$ & $45.45 / 43.31$ & $51.56 / 51.45$ \\
65536 & $58.47 / 59.44$ & $45.63 / 43.49$ & $46.97 / 44.43$ & $49.59 / 53.48$ & $45.61 / 43.62$ & $51.64 / 51.52$ \\
131072 & $58.48 / 59.42$ & $45.61 / 43.65$ & $46.89 / 44.55$ & $49.80 / 53.82$ & $45.57 / 43.72$ & $51.69 / 51.66$ \\
\bottomrule
\end{tabular}
\end{table*}

\begin{table*}[t]
\centering
\small
\setlength{\tabcolsep}{5pt}
\caption{Median raw relative $L^2$ error ($\times 10^{-2}$, lower is better) for fixed $B=8$-trained input-ablation models evaluated at budget $B$. Entries give source-conditioned / no-source results. Counts and pairing are specified in the reporting protocol.}
\label{tab:input_budget_l2}
\begin{tabular}{rcccccc}
\toprule
$B$ & Laplace & Poisson & Yukawa & Biharmonic & Helmholtz & Pooled \\
\midrule
2 & $4.24 / 4.36$ & $8.72 / 14.02$ & $4.08 / 6.54$ & $5.67 / 4.05$ & $9.94 / 14.77$ & $5.55 / 6.47$ \\
4 & $2.00 / 1.98$ & $5.55 / 10.38$ & $2.61 / 4.73$ & $4.27 / 3.05$ & $6.15 / 11.23$ & $3.07 / 3.85$ \\
8 & $1.28 / 1.28$ & $3.56 / 8.02$ & $1.77 / 3.64$ & $3.29 / 2.22$ & $4.14 / 8.53$ & $2.07 / 2.89$ \\
16 & $1.29 / 1.29$ & $2.91 / 6.45$ & $1.42 / 2.90$ & $2.51 / 1.74$ & $3.42 / 6.78$ & $1.82 / 2.37$ \\
32 & $1.34 / 1.34$ & $2.62 / 5.12$ & $1.35 / 2.42$ & $2.06 / 1.42$ & $2.81 / 5.57$ & $1.77 / 2.16$ \\
64 & $1.36 / 1.34$ & $2.33 / 4.22$ & $1.28 / 2.11$ & $1.72 / 1.21$ & $2.54 / 4.69$ & $1.66 / 1.98$ \\
128 & $1.33 / 1.30$ & $2.09 / 3.64$ & $1.25 / 1.95$ & $1.45 / 1.05$ & $2.36 / 3.97$ & $1.55 / 1.83$ \\
256 & $1.34 / 1.27$ & $1.98 / 3.31$ & $1.20 / 1.84$ & $1.30 / 0.92$ & $2.16 / 3.55$ & $1.47 / 1.67$ \\
512 & $1.35 / 1.27$ & $1.81 / 2.93$ & $1.25 / 1.81$ & $1.20 / 0.83$ & $2.09 / 3.18$ & $1.42 / 1.55$ \\
1024 & $1.36 / 1.25$ & $1.70 / 2.69$ & $1.31 / 1.74$ & $1.12 / 0.75$ & $2.05 / 2.97$ & $1.38 / 1.45$ \\
2048 & $1.35 / 1.25$ & $1.64 / 2.41$ & $1.33 / 1.76$ & $1.02 / 0.69$ & $1.90 / 2.77$ & $1.35 / 1.36$ \\
4096 & $1.36 / 1.25$ & $1.67 / 2.34$ & $1.33 / 1.76$ & $0.95 / 0.63$ & $1.84 / 2.62$ & $1.32 / 1.31$ \\
8192 & $1.35 / 1.24$ & $1.68 / 2.23$ & $1.33 / 1.75$ & $0.91 / 0.60$ & $1.83 / 2.52$ & $1.30 / 1.29$ \\
16384 & $1.36 / 1.24$ & $1.66 / 2.19$ & $1.32 / 1.74$ & $0.88 / 0.56$ & $1.78 / 2.40$ & $1.28 / 1.27$ \\
32768 & $1.36 / 1.24$ & $1.66 / 2.13$ & $1.33 / 1.73$ & $0.83 / 0.51$ & $1.77 / 2.27$ & $1.26 / 1.25$ \\
65536 & $1.36 / 1.24$ & $1.66 / 2.12$ & $1.33 / 1.73$ & $0.80 / 0.49$ & $1.77 / 2.21$ & $1.24 / 1.24$ \\
131072 & $1.36 / 1.24$ & $1.72 / 2.12$ & $1.34 / 1.73$ & $0.78 / 0.48$ & $1.78 / 2.20$ & $1.23 / 1.23$ \\
\bottomrule
\end{tabular}
\end{table*}

\subsection{Physical diagnostics}\label{app:physics_metrics}
On 300 cases (60 per family), with B=8 and no TTA, raw/MC$^2$ median relative $L^2$ is $0.8275/0.015$ and relative $H^1$ is $24.1/0.060$. Central differences use a one-pixel-eroded interior. Family-specific finite-difference residuals (five-point Laplacian and 13-point biharmonic stencil), divided by the analytic grid's discrete residual, are $1.37\times10^6/1.3\times10^3$. Median normalized two-pixel boundary-shell errors are $0.021/0.026$ (analytic grid $0.017$). Correction therefore improves derivatives/residuals but slightly worsens boundary-shell error.

\subsection{Disjoint-atom tests}\label{app:unseen_atoms}
On 500 disjoint-atom cases (100 per family), without retraining and with B=8/no TTA, median PSNR is raw $14.92$, NLM $26.70$, full-input FNO/PINO $7.27/5.70$, matched U-Net $35.46$, and MC$^2$ $42.32$ dB.

\subsection{Resolution transfer}\label{app:resolution_transfer}
On 510 test cases, the 256-trained model transfers without retraining via $256^2$ tiles, overlap 32/Hann blending and no TTA. Median raw/corrected relative $L^2$ is $1.079/0.0189$ at $256^2$ and $1.088/0.0166$ at $1024^2$ with forward times being $0.03/0.29$s.

\subsection{Point queries}\label{app:point_queries}

For 1,024 queries on each of 500 cases, MC$^2$ uses a full corrected $256^2$ grid, mask-aware bilinear interpolation and nearest-valid fallback. Median relative $L^2$/seconds is $0.019/0.88$, including fresh-grid WoS/inference, and direct pointwise WoS gives $0.230/0.78$ (B128), $0.114/0.92$ (B512), and $0.056/2.74$ (B2048). These are measured operating points, not exact equal-time comparisons or evidence against FEM.

\subsection{Neural Walk-on-Spheres}\label{app:nwos}
The median relative $L^2$/total seconds is NWoS $0.263/376$ versus MC$^2$ B8/no-TTA $0.0244/1.56$. NWoS trains per instance so we only tested on Poisson cases.

\subsection{Repeated-noise diagnostics}\label{app:empirical_bias_variance}
The repeated-noise experiment uses 300 fixed instances (60 per family), with ten independent $B=8$ realizations per instance. Appendix~\ref{app:empirical_bv_analysis} and Table~\ref{tab:empirical_bias_variance} report the full raw/corrected bias--variance comparison. The magnitude of the mean spatial error autocorrelation is at most $3\times10^{-4}$ at lags 1, 2, 4, and 8 pixels, and the median variance-field coefficient of variation is 1.26. Feeding exact solutions through the corrector gives median relative $L^2$ error $0.0043$. In a separate 50-instance sensitivity diagnostic, the normalized noisy-channel Jacobian norm has median $119.6$, while the raw-unit finite-difference gain between actual seed pairs has median $0.0118$. These quantities use different coordinate systems and do not establish a global contraction bound.

\subsection{Budget dependence}\label{app:budget_sweep}
At B=8 all test cases improve raw masked MSE and raw/corrected median MSE is $0.906/2.41\times10^{-4}$. At B=32,768 it is $2.42\times10^{-4}/1.65\times10^{-5}$ with $92.85\%$ improved ($n=4,949$). At B=131,072, $6.68\times10^{-5}/1.56\times10^{-5}$ with $76.65\%$ improved ($n=4,946$).

\subsection{Compute accounting}\label{app:compute}
We time 50 problems (10 per PDE family) at $256\times256$ resolution and batch size one on A100-SXM4-80GB GPUs. Four workers each use a dedicated GPU. WoS uses a 256-step outer cap, a 16-step biharmonic inner cap, and $1024\times1024$ source grids for biharmonic and Helmholtz problems. Times are cumulative within one continuous WoS run, rather than independent solves at each budget. Neural inference is timed after warm-up with GPU synchronization. Table~\ref{tab:compute_budget_ladder} excludes model loading, warm-up, input normalization and preparation, associated transfers, and file I/O; separately timed stages appear in Table~\ref{tab:compute_stages}.

\begin{table}[t]
\centering
\small
\caption{Median time in seconds over 50 cases. Combined times are medians of per-case sums, using the neural overhead measured at $B=8$ for every budget. TTA denotes eight-pass rotation/reflection augmentation; headline results use a single pass.}
\label{tab:compute_budget_ladder}
\begin{tabular}{rrrr}
\toprule
$B$ & WoS & WoS + single pass & WoS + TTA \\
\midrule
1 & 1.777 & 1.806 & 2.011 \\
2 & 1.778 & 1.807 & 2.012 \\
4 & 1.778 & 1.807 & 2.012 \\
8 & 1.778 & 1.807 & 2.012 \\
16 & 1.778 & 1.808 & 2.013 \\
32 & 1.779 & 1.808 & 2.013 \\
64 & 1.779 & 1.808 & 2.013 \\
128 & 1.779 & 1.809 & 2.013 \\
256 & 1.780 & 1.809 & 2.014 \\
512 & 3.353 & 3.382 & 3.586 \\
1024 & 6.315 & 6.344 & 6.546 \\
2048 & 12.561 & 12.590 & 12.792 \\
4096 & 25.169 & 25.198 & 25.400 \\
8192 & 50.239 & 50.268 & 50.470 \\
16384 & 100.319 & 100.347 & 100.549 \\
32768 & 200.518 & 200.547 & 200.749 \\
65536 & 400.768 & 400.797 & 400.999 \\
131072 & 801.199 & 801.228 & 801.430 \\
\bottomrule
\end{tabular}
\end{table}

\begin{table}[t]
\centering
\small
\caption{Median time in seconds over 50 cases. All budgets are measured within one continuous WoS run, so each entry is the time needed to reach $B$ walks per point. The solver processes walks in batches of 256, so every budget up to $B=256$ is completed by the first batch and shares its time ($\approx 1.78$\,s); these entries overstate the cost of very small budgets, and a standalone $B=8$ solve takes $0.891$\,s. Beyond $B=256$, time grows linearly with $B$. Combined times add the neural overhead measured at $B=8$ (median of per-case sums). TTA denotes eight-pass rotation/reflection augmentation; headline results use a single pass.}
\label{tab:compute_stages}
\begin{tabular}{lr}
\toprule
Stage & Time (ms) \\
\midrule
Input loading and function construction & 278.61 \\
Single-pass neural inference & 28.95 \\
Eight-pass augmented inference & 231.78 \\
Output transfer and serialization & 23.81 \\
\bottomrule
\end{tabular}
\end{table}

A separate 20-case measurement of independently launched WoS solves gives $0.891$\,s at $B=8$ and $2.412$\,s at $B=256$. These standalone timings are not combined with the cumulative measurements to calculate speedups.

The headline SUNet, matched FNO, and matched U-Net use $99.681/99.799/99.560$ million parameters, $64.03/15.57/352.89$ GFLOPs per forward, and $685/711/690$ MiB peak memory, respectively. Online timings exclude data generation and training, whose amortized cost over $N$ solves is $(T_{\mathrm{data}}+T_{\mathrm{train}})/N$. The recorded $10.35$ hours on four A100 GPUs covers only the final training segment, not total offline cost.

\paragraph{\textcolor{black}{Diffusion inference timing.}} \textcolor{black}{A separate benchmark measures the model stage on one A100-SXM4-80GB at batch size one, after warm-up and GPU synchronization. Uncompiled MC$^2$ takes a median $0.0665$\,s with IQR $0.0665$--$0.0666$\,s. The 100-step DiffusionPDE-mode sampler takes $11.37$\,s with IQR $11.28$--$11.46$\,s when uncompiled, or $6.81$\,s with IQR $6.73$--$7.01$\,s with \texttt{torch.compile}. Relative to MC$^2$ in this benchmark, the ratios of median model-stage times are $171\times$ and $102\times$, respectively.} \par \subsection{Normalization audit}\label{app:normalization_audit}

Historical model inputs pool noisy-field and target extrema, exposing target information. This differs from GT ranges used only for metrics. Holding checkpoints fixed, the completed input-only audit changes median raw MC$^2$ relative $L^2$ by at most $0.11\%$ at every budget across test cases.

\subsection{Geometry and failures}\label{app:failure_modes}

With shared weights, $B=8$, and single-pass correction without test-time augmentation, MC$^2$ improves median PSNR by $28$–$34$ dB on every domain type. Boolean-composed domains ($n=1321$) are not the hardest: raw WoS/MC$^2$ median PSNR is $16.03/47.19$ dB, comparable to annuli ($18.60/48.21$ dB, $n=442$), stadiums ($17.27/48.36$ dB, $n=448$), and triangles ($17.40/48.70$ dB, $n=457$). The lowest corrected values occur on simple primitives whose raw estimates are noisiest: squares ($10.59/38.55$ dB, $n=457$), disks ($9.05/40.04$ dB, $n=441$), and rectangles ($10.11/41.34$ dB, $n=471$), with ellipses ($10.79/44.31$ dB) and hexagons ($12.04/45.05$ dB) in between. Corrected quality therefore tracks the noise level of the input rather than geometric complexity.


\section{Comparison with Finite Element Methods}
\label{app:fem}
\label{app:fem_results}
\subsection{Accelerating Monte Carlo without volumetric meshing}
\label{app:fem_scope}

MC$^2$ learns to correct the output of a Monte Carlo solver, allowing a low-budget estimate to recover accuracy that would otherwise require substantially more sampling. The error-correction framework preserves the underlying solver's representation of the domain. Applying the learned corrector does not require constructing a body-fitted volumetric mesh. The practical question is whether this combination of low-budget WoS estimates and learned error correction is competitive with a deterministic solve, and how that comparison changes with geometry and dimension.

Our main experiments compare WoS, classical denoising methods, and learned baselines. Like WoS, these methods do not require a body-fitted mesh. Finite element methods (FEM) also require mesh construction, whose cost and robustness affect the comparison \citep{Sawhney:2020:MCG}. We therefore evaluate FEM separately here, including mesh construction costs. We first establish the accuracy--compute tradeoff on the dense 2D benchmark, where FEM performs particularly well. We then extend MC$^2$ to 3D and evaluate both accuracy and the ability to produce an estimate on analytic, real-world, and implicit geometries. These experiments show where Monte Carlo error correction provides a practical benefit. MC$^2$ remains effective when volumetric meshing is unreliable, and offers favorable accuracy--compute tradeoffs on the tested 3D analytic domains. Higher dimension or geometric complexity alone does not guarantee this advantage, and we also report settings where FEM remains preferable.

\paragraph{Comparison protocol.}
FEM uses quadratic (P2) elements on body-fitted meshes, implemented with Triangle \citep{shewchuk1996triangle}, gmsh \citep{geuzaine2009gmsh}, and scikit-fem \citep{gustafsson2020scikitfem}. Both numerical solvers receive the domain, source term, and boundary conditions, including the additional boundary conditions on the Laplacian for the biharmonic problems. MC$^2$ additionally uses a pretrained neural corrector. FEM timings include mesh construction, assembly and solve, and evaluation of the resulting field, on one CPU core. MC$^2$ timings include generation of its WoS input and network inference on one A100. Offline training is excluded. Unless stated otherwise, errors are median relative $L^2$ errors against analytic ground truth. These are comparisons of the evaluated implementations and hardware configurations.

\subsection{Accuracy, runtime, and mesh refinement in two dimensions}
\label{app:fem_2d}

On the evaluated PDEZoo test problems, FEM achieves both lower median error and lower median runtime than MC$^2$ at $B=8$ (Table~\ref{tab:fem_2d}). At $h=0.08$, FEM is more accurate on approximately $99\%$ of problems, with per-family win rates of approximately $97$--$100\%$. Refining the mesh further improves accuracy. Thus, accelerating WoS does not by itself make MC$^2$ the preferred solver for dense evaluation on these 2D domains.

\begin{table}[t]
\centering
\small
\begin{tabular}{lcc}
\toprule
Method & Median relative $L^2$ & Median wall-clock time (s) \\
\midrule
MC$^2$, $B=8$ & $0.018$ & $\sim 1.8$ \\
FEM, $h=0.08$ & $6.6\times10^{-4}$ & $0.73$ \\
FEM, $h=0.04$ & $9.7\times10^{-5}$ & $1.2$ \\
FEM, $h=0.02$ & $1.3\times10^{-5}$ & $4.9$ \\
\bottomrule
\end{tabular}
\caption{Dense 2D evaluation on 5,000 test problems. FEM timings include meshing, solving, and field evaluation on one CPU core, while MC$^2$ includes WoS and correction on one A100. Error and time are separate medians over problems.}
\label{tab:fem_2d}
\end{table}

\paragraph{Increasing geometric complexity.}
We also evaluate 20 perforated-plate problems across all five PDE families, with 20--200 circular holes occupying approximately $25\%$ of the area. FEM at $h=0.08$ is more accurate on 19 of 20 problems. At 200 holes, the median mesh contains approximately 36,000 triangles, with median error $8.8\times10^{-5}$ and runtime $2.46$\,s. MC$^2$ reaches $9.9\times10^{-3}$ in $1.78$\,s. The tested planar mesher therefore handles substantial perforation efficiently. The relevant distinction is the difficulty of constructing a suitable mesh, rather than the number of geometric features alone.

\paragraph{Finite-compute correction and the error floor.}
MC$^2$ provides its largest benefit when the input is dominated by Monte Carlo error. Its median 2D error decreases from $0.018$ at $B=8$ to $0.0082$ at $B=128$, but approaches a floor near $5\times10^{-3}$ at larger budgets ($0.0051$ at $B=131{,}072$). FEM continues to improve over the tested mesh refinements. This behavior highlights the finite-compute setting of MC$^2$. Learned correction substantially reduces sampling error, but its remaining correction error limits the accuracy attainable by increasing the WoS input budget alone.

\subsection{Retaining the benefits of a mesh-free solver in three dimensions}
\label{app:fem_3d}

The error-correction framework extends directly to three-dimensional solution fields. We train a separate 3D residual U-Net with 11.8M parameters on 20,000 PDE instances on analytic domains at resolution $64^3$, using WoS budgets $B\in\{1,2,4,8,16,32\}$. We then fine-tune on a mixture including 1,600 Thingi10K training models disjoint from the test models. All MC$^2$ results below use this fine-tuned model and $B=8$ inputs. This experiment evaluates a 3D implementation of the framework, rather than transfer of the trained 2D network.

Evaluation includes 500 PDE instances on analytic domains, 100 Thingi10K \citep{zhou2016thingi10k} models (50 clean and 50 defective, with five PDE problems per model), and 100 implicit domains (gyroid lattices \citep{schoen1970gyroid} and smoothly blended blobs, with four problems per domain). The implicit set excludes biharmonic problems because of the cost of nested WoS walks. All test sets, the seeded and stratified Thingi10K selection, and this exclusion were fixed before any results were computed. The 3D FEM pipeline uses gmsh tetrahedral meshing, with a 300\,s meshing limit. For implicit domains, its timing includes surface extraction. Meshing errors and timeouts count as unsolved problems.

\paragraph{Accuracy and runtime on analytic domains.}
On the analytic test set, MC$^2$ reaches median relative $L^2$ error $0.0100$ in $1.8$\,s, compared with $0.3912$ for its $B=8$ WoS input and $0.0982$ for WoS at $B=128$ in $13.8$\,s. Thus, correction gives approximately $9.8\times$ lower median error than using $16\times$ as many walks. FEM spans errors from $0.0203$ at $h=0.2$ in $3.8$\,s to $0.0039$ at $h=0.1$ in $10.5$\,s and $0.0014$ at $h=0.07$ in $14.8$\,s. MC$^2$ therefore offers a useful accuracy--compute tradeoff, with substantially lower wall-clock time than FEM at the finer mesh resolutions.

We additionally estimate the FEM time required to match each problem's MC$^2$ error. The median ratio of estimated FEM time to MC$^2$ time is $5.14\times$ (95\% bootstrap interval $[4.31,6.08]$), with MC$^2$ faster on 395 of 500 problems under this procedure. We interpolate FEM error--time curves in log space on 371 problems and extrapolate beyond the finest tested mesh on 70. On the remaining 59, the coarsest tested FEM mesh already achieves lower error, so its runtime gives an upper bound on the time needed to match MC$^2$. This estimate supplements the directly measured operating points, but is not an exact equal-error timing measurement on every problem.

\paragraph{Geometries with unreliable volumetric meshing.}
MC$^2$ and WoS produce estimates for every tested problem. In contrast, at $h=0.1$ the gmsh-based FEM pipeline completes 160 of 250 clean Thingi10K problems and 10 of 250 defective problems. On gyroid lattices, it completes 72 of 200 at $h=0.2$ and 116 of 200 at $h=0.1$. The corresponding failure or timeout rates are $36\%$, $96\%$, and $42$--$64\%$, respectively. These results demonstrate that the error-correction framework improves low-budget WoS solutions without requiring a volumetric mesh.

\begin{table}[t]
\centering
\small
\resizebox{\linewidth}{!}{%
\begin{tabular}{lccccc}
\toprule
Geometry & FEM completed & MC$^2$ completed & FEM error (time) & MC$^2$ error (time) & WoS $B=128$ error \\
\midrule
Analytic domains & $500/500$ & $500/500$ & $0.0039$ ($10.5$\,s) & $0.0100$ ($1.8$\,s) & $0.0982$ \\
Thingi10K, clean & $160/250$ & $250/250$ & $3\times10^{-4}$ ($23.4$\,s) & $0.0120$ ($4.2$\,s) & $0.0224$ \\
Thingi10K, defective & $10/250$ & $250/250$ & $4\times10^{-4}$ ($11.7$\,s) & $0.0125$ ($5.9$\,s) & $0.0272$ \\
Gyroid lattices & $116/200$ & $200/200$ & $0.0088$ ($11.2$\,s) & $0.0117$ ($5.0$\,s) & $0.0512$ \\
Blended blobs & $200/200$ & $200/200$ & $0.0029$ ($6.4$\,s) & $0.0125$ ($13.4$\,s) & $0.0714$ \\
\bottomrule
\end{tabular}%
}
\caption{Completion, accuracy, and runtime in 3D. FEM completion counts, errors, and times use $h=0.1$ throughout. FEM errors and times are medians over completed problems only. MC$^2$ and WoS complete all problems, and their errors and times are medians over each full test set. MC$^2$ uses $B=8$, giving approximately $1.9$--$9.8\times$ lower median error than WoS with $16\times$ the WoS budget. Counts are PDE problems, with multiple problems per geometry. Completion measures the ability to return an estimate, not satisfaction of an accuracy tolerance.}
\label{tab:fem_3d}
\end{table}

\paragraph{Where FEM remains preferable.}
The 3D advantage depends on the geometry and accuracy requirement. On clean Thingi10K problems successfully solved at $h=0.1$, FEM reaches median error approximately $3\times10^{-4}$. In the paired comparison at matched error, the coarsest available FEM result already beats MC$^2$ on 141 of these 160 problems. On smooth blended blobs, FEM succeeds on all 200 problems and reaches median error $0.0029$ in $6.4$\,s, compared with $0.0125$ in $13.4$\,s for MC$^2$. Three dimensions alone therefore does not imply an advantage over FEM. The evidence supports favorable runtime on the tested analytic domains and reliable production of estimates on geometries where this volumetric-meshing pipeline frequently fails.

\subsection{Validation against independent FEM references}
\label{app:fem_independent}

To assess correction beyond evaluation against analytic ground truth, we separately construct 500 problems with independently specified source terms and boundary conditions, covering 100 problems per family and nine domain types. We solve these problems with P2 FEM to obtain reference fields. Reference refinement uses an error estimate equal to one eighth of the relative change between successive meshes, with a target tolerance of $10^{-4}$. All 500 references satisfy this tolerance, with median estimated error $5.8\times10^{-8}$. This is a mesh-refinement estimate, distinct from the FEM solver's separate validation against analytic solutions. Degenerate sampled domains are rejected before solving. Accepted domains must have at least 1,000 interior grid points, median interior distance to the boundary at least $0.03$, and nonempty FEM geometry.

Against these independent references, MC$^2$ achieves median relative $L^2$ error $0.0047$ (95\% percentile bootstrap interval $[0.00430,0.00505]$, using 2,000 resamples of the 500 problems), compared with $0.207$ for $B=8$ WoS, $0.036$ for non-local means (NLM), and approximately $0.023$ for each of the FNO and PINO denoising baselines. These FNO/PINO models take noisy WoS fields as input. They are distinct from the neural-operator baselines that predict the solution without a WoS input. This experiment supports the effectiveness of Monte Carlo error correction when the evaluation reference is produced by an independent numerical solver.

\subsection{Limitations of the comparison}

\label{app:fem_limitations}
The observed 3D failures belong to the tested gmsh-based pipeline, rather than FEM as a numerical method. Alternative tetrahedralizers were not evaluated and may change both completion rates and runtime. Timing comparisons use a single-core Python FEM implementation, with WoS and neural correction running on a GPU. Optimized or parallel FEM implementations could change the tradeoff. Although WoS accesses geometry through its geometric queries, the learned correction operates on a fixed $64^3$ grid, which limits representation of thin features. The 3D results use one training seed, and the reported bootstrap interval measures variation over test problems rather than training runs. Biharmonic coverage is limited in fine-tuning and absent from the implicit-domain set. Within these limits, the experiments show that MC$^2$ can extend the useful finite-compute regime of WoS while retaining its ability to operate without a body-fitted volumetric mesh.

\section{Bias--Variance Analysis of PDE Solvers}
\label{app:bias_variance}

Here, we formalize the bias--variance tradeoff among three estimators of an  elliptic PDE solution $u$: the pure Monte Carlo Walk-on-Spheres estimator $\hat u_K$, a pure neural operator $\Phi_\theta^{\mathrm{op}}$, and our 
hybrid MC$^2$ estimator $\tilde u_K = \Phi_\theta(\hat u_K, a)$, where $a$ denotes any auxiliary conditioning information available to the  corrector, such as the source term, domain mask, boundary data, or PDE 
parameters. Throughout, we use the standard bias--variance decomposition of mean-squared error under squared loss \citep{geman1992neural,hastie2009elements}.  For MC$^2$, $a=f$. The conditional-mean results below concern an idealized squared-loss population objective, whereas the checkpoint uses masked $\ell_1$ training. Assume finite second moments wherever variances are used.
We present the analysis pointwise at a fixed query $x\in\Omega$, while the same conclusions hold for discretized fields by summing over grid points, or in $L^2(\Omega)$ by integration. For an estimator $T$ of $u(x)$, the conditional mean-squared error decomposes as
\begin{equation}
\mathbb{E}\!\left[(T-u(x))^2 \mid P\right]
=
\underbrace{
\left(\mathbb{E}[T\mid P]-u(x)\right)^2
}_{\mathrm{Bias}^2(T\mid P)}
+
\underbrace{
\mathrm{Var}(T\mid P)
}_{\mathrm{Variance}(T\mid P)},
\label{eq:bv-decomp}
\end{equation}
where $P$ denotes the underlying PDE instance and the expectation is over the 
internal randomness of the estimator.

\subsection{Monte Carlo WoS Solver}

Let $\hat u_K(x)=K^{-1}\sum_{k=1}^K Y_k(x)$ average independent identically distributed complete walk payoffs, with fixed numerical settings $\eta$ suppressed in the notation. WoS expresses  the solution of elliptic boundary-value problems as the expectation of a random  variable induced by Green's functions and boundary hitting distributions  \citep{Muller,Sawhney:2020:MCG}. For these numerical settings, define the combined bias $b_\eta(P,x)$ by
\begin{equation}
\mathbb E[\hat u_K(x)\mid P]=u(x)+b_\eta(P,x).
\end{equation}

This bias includes shell termination, finite-cap projection, and interpolation, and need not vanish.

The variance of the average of $K$ independent walks is
\begin{equation}
    \mathrm{Var}(\hat u_K(x)\mid P)
    =
    \frac{\sigma_P^2(x)}{K},
\end{equation}
where $\sigma_P^2(x)=\mathrm{Var}(Y_1(x)\mid P)$ is the single-walk variance. 
Therefore,
\begin{equation}
\mathbb{E}\!\left[
    (\hat u_K(x)-u(x))^2 \mid P
    \right]
    =
    \frac{\sigma_P^2(x)}{K}
    +
b_\eta(P,x)^2.
\label{eq:wos-mse}
\end{equation}

Thus variance decreases as $1/K$, while fixed numerical bias can produce an error floor.

\subsection{Neural Operators}

A pure neural operator maps a PDE specification directly to a solution field, $\Phi_\theta^{\mathrm{op}} : (f,g,\Omega) \mapsto \hat u^{\mathrm{op}}$, as in neural operator and physics-informed neural operator methods 
\citep{li2021fourier,DBLP:journals/corr/abs-2111-03794}. 
Conditional on a PDE instance $P$, the prediction is deterministic at inference 
time, so
\begin{equation}
    \mathrm{Var}(\Phi_\theta^{\mathrm{op}}(P)(x)\mid P)=0.
\end{equation}
However, the estimator generally has approximation bias
\begin{equation}
    b_\theta(P,x)
    =
    \Phi_\theta^{\mathrm{op}}(P)(x)-u(x),
\end{equation}
so its conditional MSE is
\begin{equation}
    \mathbb{E}\!\left[
    \left(\Phi_\theta^{\mathrm{op}}(P)(x)-u(x)\right)^2
    \mid P
    \right]
    =
    b_\theta(P,x)^2.
\end{equation}
Unlike Monte Carlo error, this bias does not decrease with additional 
inference-time sampling compute. Moreover, distribution shift can increase learned approximation error, while numerical WoS has the separate approximation bias described above \citep{10.1145/3610548.3618141,cuomo2022scientific}.

\subsection{MC$^2$: Learned Correction of a Monte Carlo Estimator}

MC$^2$ takes a finite-budget Monte Carlo estimate and optional auxiliary PDE 
information as input. Let $X_K = (\hat u_K, a)$, where $a$ may include the source term, domain mask, boundary information, or  PDE parameters. MC$^2$ outputs $\tilde u_K = \Phi_\theta(X_K)$. The population-optimal corrector under squared loss is the conditional  expectation, also known as the regression function or Bayes estimator under 
squared loss \citep{hastie2009elements}:
\begin{equation}
    \Phi^\star(X_K)(x)
    =
    \mathbb{E}\!\left[u(x)\mid X_K\right].
\label{eq:bayes-corrector}
\end{equation}
Equivalently,
\begin{equation}
    \Phi^\star
    =
    \arg\min_{\Phi}
    \mathbb{E}
    \left[
    \|\Phi(X_K)-u\|_2^2
    \right].
\end{equation}

This immediately gives a useful comparison to raw WoS. Since the identity map 
$\Phi(X_K)=\hat u_K$ is an admissible candidate in the minimization above, 
the Bayes-optimal corrector is never worse than raw WoS in population MSE:
\begin{equation}
    \mathbb{E}
    \left[
    \|\Phi^\star(X_K)-u\|_2^2
    \right]
    \leq
    \mathbb{E}
    \left[
    \|\hat u_K-u\|_2^2
    \right].
\label{eq:bayes-better-wos}
\end{equation}
If $X_K$ also contains every input that a standalone neural operator would 
use, such that $\Phi_\theta^{\mathrm{op}}(P)$ can be expressed as a  measurable function of $X_K$, then the operator is also an admissible candidate in the minimization, and
\begin{equation}
    \mathbb{E}
    \left[
    \|\Phi^\star(X_K)-u\|_2^2
    \right]
    \leq
    \mathbb{E}
    \left[
    \|\Phi_\theta^{\mathrm{op}}(P)-u\|_2^2
    \right].
\label{eq:bayes-better-op}
\end{equation}
Thus, the optimal corrector dominates both raw Monte Carlo and standalone 
neural prediction whenever the same information is available. For a square-integrable learned predictor, conditional-expectation orthogonality gives the following excess-risk identity, whose nonnegative final term does not guarantee improvement:
\begin{equation}
    \mathbb{E}
    \left[
    \|\Phi_\theta(X_K)-u\|_2^2
    \right]
    =
    \mathbb{E}
    \left[
    \|\Phi^\star(X_K)-u\|_2^2
    \right]
    +
    \mathcal{E}_{\mathrm{approx}}(\theta),
\label{eq:approx-error}
\end{equation}
where $\mathcal{E}_{\mathrm{approx}}(\theta)\geq 0$ captures the gap between 
the learned corrector and the Bayes-optimal corrector.

\subsection{Empirical Bias--Variance Analysis}
\label{app:empirical_bv_analysis}

For the trained MC$^2$ corrector, we measure bias and variance directly. At a fixed PDE instance, let $V$ denote conditional variance averaged over valid pixels and $b^2$ the corresponding mean squared bias. By Eq.~\eqref{eq:bv-decomp}, correction improves MSE exactly when
\begin{equation}
V_{\mathrm{raw}}-V_{\mathrm{corr}}
>
b_{\mathrm{corr}}^2-b_{\mathrm{raw}}^2.
\label{eq:mc2-improvement-condition}
\end{equation}
This condition does not require the corrector to be contractive.

We use the repeated-noise experiment in Appendix~\ref{app:empirical_bias_variance}: 300 fixed PDE instances, with 60 per family and $S=10$ independent WoS realizations at $B=8$ per instance. MC$^2$ uses one forward pass without test-time augmentation. Measurements use raw fields without metric clipping, with the input-normalization protocol discussed in Appendix~\ref{app:normalization_audit}. For either estimator $T$, we compute
\[
\bar T=\frac{1}{S}\sum_{s=1}^{S}T_s,\qquad
\widehat V=\frac{1}{S-1}\sum_{s=1}^{S}\|T_s-\bar T\|_V^2,\qquad
\widehat{b^2}=\|\bar T-u\|_V^2-\frac{\widehat V}{S},
\]
Here $\|v\|_V^2$ averages $v^2$ over valid pixels. The finite-sample-corrected squared-bias estimate is signed and can be negative. Table~\ref{tab:empirical_bias_variance} reports the estimates.

\begin{table}[!htbp]
\centering
\small
\caption{Bias--variance estimates at $B=8$ over 300 instances with 10 realizations each. All quantities use squared raw field units, averaged over valid pixels and then aggregated over instances as indicated.}
\label{tab:empirical_bias_variance}
\begin{tabular}{lrr}
\toprule
Statistic & Raw WoS & MC$^2$ \\
\midrule
Mean variance & $351.2$ & $0.0646$ \\
Mean estimated squared bias & $0.33$ & $0.0269$ \\
Median estimated squared bias & $7.5\times10^{-7}$ & $5.8\times10^{-5}$ \\
Mean MSE & $351.5$ & $0.0916$ \\
\bottomrule
\end{tabular}
\end{table}

The ratio of mean MSEs is approximately $3{,}800$. The mean per-instance corrected/raw variance ratio is $2.21\times10^{-4}$. These results demonstrate empirical variance reduction, but establish neither raw-WoS unbiasedness nor global contraction. Appendix~\ref{app:empirical_bias_variance} gives further spatial-noise and sensitivity diagnostics.

\subsection{Budget Dependence and High-Budget Limit}

At $B=131{,}072$, median raw and corrected MSE are $6.68\times10^{-5}$ and $1.56\times10^{-5}$, respectively, a ratio of approximately $4.3$. Correction improves $76.65\%$ of 4,946 cases (Appendix~\ref{app:budget_sweep}). On exact solution inputs, median relative $L^2$ error is $0.0043$ (Appendix~\ref{app:empirical_bias_variance}), measuring deviation from the identity.

For fixed numerical settings $\eta$ and integrable payoffs, the law of large numbers gives $\hat u_K\to u+b_\eta$. If the complete correction map is continuous at this limit,
\begin{equation}
\tilde u_K=\Phi_\theta(\hat u_K,f)
\to\Phi_\theta(u+b_\eta,f).
\end{equation}
Consistency therefore requires $\Phi_\theta(u+b_\eta,f)=u$, which is not established. The clean-input diagnostic evaluates the map at $u$, rather than at $u+b_\eta$, so it does not by itself determine this high-budget limit. Our evidence supports finite-budget error reduction without a guarantee of asymptotic consistency.


\newpage

\clearpage
\section*{NeurIPS Paper Checklist}


\begin{enumerate}

\item {\bf Claims}
    \item[] Question: Do the main claims made in the abstract and introduction accurately reflect the paper's contributions and scope?
    \item[] Answer: \answerYes{} 
    \item[] Justification: The major contributions in the abstract and introduction, MC$^2$ and PDEzoo are supported in the main body with detailed analysis, comparison to other SOTA benchmarks/baselines, and ample discussion. {}

\item {\bf Limitations}
    \item[] Question: Does the paper discuss the limitations of the work performed by the authors?
    \item[] Answer: \answerYes 
    \item[] Justification: We discuss limitations in the conclusion, and in particular about how $MC^2$ is only applicable to what WoS and other monte carlo methods can solve.

\item {\bf Theory assumptions and proofs}
    \item[] Question: For each theoretical result, does the paper provide the full set of assumptions and a complete (and correct) proof?
    \item[] Answer: \answerNA{} 
    \item[] Justification: Our paper is primarily not theoretical, but we do provide some proofs of bounded bias in the appendix. 

    \item {\bf Experimental result reproducibility}
    \item[] Question: Does the paper fully disclose all the information needed to reproduce the main experimental results of the paper to the extent that it affects the main claims and/or conclusions of the paper (regardless of whether the code and data are provided or not)?
    \item[] Answer: \answerYes{} 
    \item[] Justification: We give the exact architecture and training regime in the appendix, and provide a detailed overview in the main body. 

\item {\bf Open access to data and code}
    \item[] Question: Does the paper provide open access to the data and code, with sufficient instructions to faithfully reproduce the main experimental results, as described in supplemental material?
    \item[] Answer: \answerYes{} 
    \item[] Justification: The provide code, dataset, the dataset generator, and $MC^2$ model weights fully open-source.

\item {\bf Experimental setting/details}
    \item[] Question: Does the paper specify all the training and test details (e.g., data splits, hyperparameters, how they were chosen, type of optimizer) necessary to understand the results?
    \item[] Answer: \answerYes{} 
    \item[] Justification: A full description is provided in the appendix

\item {\bf Experiment statistical significance}
    \item[] Question: Does the paper report error bars suitably and correctly defined or other appropriate information about the statistical significance of the experiments?
    \item[] Answer: \answerYes{} 
    \item[] Justification: We provide standard deviations in our tables containing the results of our experiments.

\item {\bf Experiments compute resources}
    \item[] Question: For each experiment, does the paper provide sufficient information on the computer resources (type of compute workers, memory, time of execution) needed to reproduce the experiments?
    \item[] Answer: \answerYes{} 
    \item[] Justification: We specifically address this in the appendix 
    
\item {\bf Code of ethics}
    \item[] Question: Does the research conducted in the paper conform, in every respect, with the NeurIPS Code of Ethics \url{https://neurips.cc/public/EthicsGuidelines}?
    \item[] Answer: \answerYes{} 
    \item[] Justification: We conform to each and every part of the ethics guidelines, and the contents presented in our paper does not pose any direct or foreseeable direct threat to safety. We also respect data/privacy guidelines and follow the impact mitigation measures.

\item {\bf Broader impacts}
    \item[] Question: Does the paper discuss both potential positive societal impacts and negative societal impacts of the work performed?
    \item[] Answer: \answerNA{} 
    \item[] Justification: Our paper does not have any societal impact.
    
\item {\bf Safeguards}
    \item[] Question: Does the paper describe safeguards that have been put in place for responsible release of data or models that have a high risk for misuse (e.g., pre-trained language models, image generators, or scraped datasets)?
    \item[] Answer: \answerNA{} 
    \item[] Justification: Our paper does not pose such risk. 

\item {\bf Licenses for existing assets}
    \item[] Question: Are the creators or original owners of assets (e.g., code, data, models), used in the paper, properly credited and are the license and terms of use explicitly mentioned and properly respected?
    \item[] Answer: \answerYes{} 
    \item[] Justification: We any work we relied on on for MC$^2$ (e.g SUNet and WoS) alongside many advancement works in the field by which we drew inspiration.

\item {\bf New assets}
    \item[] Question: Are new assets introduced in the paper well documented and is the documentation provided alongside the assets?
    \item[] Answer: \answerYes{} 
    \item[] Justification: We release a new dataset for elliptic PDE solving and it is well documented. We also create several visuals and diagrams, all of which have descriptive captions and discussion within the paper. We do not use assets that are not open source.

\item {\bf Crowdsourcing and research with human subjects}
    \item[] Question: For crowdsourcing experiments and research with human subjects, does the paper include the full text of instructions given to participants and screenshots, if applicable, as well as details about compensation (if any)? 
    \item[] Answer: \answerNA{} 
    \item[] Justification: N/A

\item {\bf Institutional review board (IRB) approvals or equivalent for research with human subjects}
    \item[] Question: Does the paper describe potential risks incurred by study participants, whether such risks were disclosed to the subjects, and whether Institutional Review Board (IRB) approvals (or an equivalent approval/review based on the requirements of your country or institution) were obtained?
    \item[] Answer: \answerNA{} 
    \item[] Justification: We do not use crowdsourcing nor does our paper involve human subjects.

\item {\bf Declaration of LLM usage}
    \item[] Question: Does the paper describe the usage of LLMs if it is an important, original, or non-standard component of the core methods in this research? Note that if the LLM is used only for writing, editing, or formatting purposes and does \emph{not} impact the core methodology, scientific rigor, or originality of the research, declaration is not required.
    \item[] Answer: \answerNA{} 
    \item[] Justification: We only use LLMs for "writing, editing, or formatting purposes" and it does not impact the core methodology or scientfiic rigor.

\end{enumerate}

\end{document}